# A cognitive neural architecture able to learn and communicate through natural language

Bruno Golosio[a*], Angelo Cangelosi[b], Olesya Gamotina[a], Giovanni Luca Masala[a]

[a]POLCOMING Department, Section of Engineering and Information Technologies,

University of Sassari, Italy

[b]Centre for Robotics and Neural Systems, School of Computing and Mathematics,

University of Plymouth, United Kingdom.

**Abstract.** Communicative interactions involve a kind of procedural knowledge that is used by the human brain for processing verbal and nonverbal inputs and for language production. Although considerable work has been done on modeling human language abilities, it has been difficult to bring them together to a comprehensive *tabula rasa* system compatible with current knowledge of how verbal information is processed in the brain. This work presents a cognitive system, entirely based on a large-scale neural architecture, which was developed to shed light on the procedural knowledge involved in language elaboration. The main component of this system is the central executive, which is a supervising system that coordinates the other components of the working memory. In our model, the central executive is a neural network that takes as input the neural activation states of the short-term memory and yields as output mental actions, which control the flow of information among the working memory components through neural gating mechanisms.

The proposed system is capable of learning to communicate through natural language starting from *tabula rasa*, without any *a priori* knowledge of the structure of phrases, meaning of words, role of the different classes of words, only by interacting with a human through a text-based interface, using an open-ended incremental learning process. It is able to learn nouns, verbs, adjectives, pronouns and other word classes, and to use them in expressive language. The model was validated on a corpus of 1587 input sentences, based on literature on early language assessment, at the level of about 4-years old child, and produced 521 output sentences, expressing a broad range of language processing functionalities.

[*]corresponding author; email: golosio@uniss.it

# 1    Introduction

The attempts to build artificial systems capable of simulating important aspects of human cognitive abilities have a long history, and have contributed to the debate among two different theoretical approaches, the computationalism and the connectionism. According to the computational theory of mind, the brain is an information processing system, and thought can be described as a computation that operates on mental states [1,2]. This perspective has led to the implementation of a class of cognitive architectures called symbolic [3-5] (see Ref.s [6] and [7] for a review). Symbolic architectures can realize high-level cognitive functions, such as complex reasoning and planning. However, the main issue of such architectures is that all information must be represented and processed in the form of symbols pertaining to a predefined domain. This constraint makes it difficult for such systems to recognize regularities in large datasets, particularly in presence of noisy data and in dynamic environments. On the other hand, the central idea of the connectionist approach is that mental processes can be modeled as emergent processes of networks of highly interconnected (subsymbolic) processing units. The most used type of connectionist model is the artificial neural network (ANN) model, which has been widely used to account for different aspects of human cognition, including memory, perception, attention, pattern recognition and language. In many cases, connectionist architectures have been very effective in explaining some features of human behavior described by psychological findings. However, up to now they have never been implemented in large scale simulations for tasks that require complex reasoning [6]. Recently, Eliasmith et al. proposed a 2.5-million neuron model of the brain, able to process visual image sequences and to respond through movements of a physically modeled arm [8]. Other large-scale neural simulations have been reported [9,10], however they focus on biological realism of the neuron model, while none of them deal with the problem of natural language elaboration.

The symbolic approach dominated the research in the field of natural language processing (NLP) for several decades. Natural language itself appears to be a strong symbolic activity, because words can be considered symbols used to represent real objects, concepts, events, and actions. On the other hand, the subsymbolic approach demonstrated to be more suitable for modeling the cognitive foundations of language processing and for representing statistical regularities in natural language [11-13]. Neural network language models have widely been used in NLP, demonstrating superior performances in next-word prediction and other standard NLP tasks over conventional approaches, such as n-gram models. Recently, deep learning techniques based on recurrent neural networks (RNNs) have been used successfully for several NLP tasks, including speech recognition [14], parsing [15,16], machine translation [17], sentiment analysis of text [18]. Although some of these models are biologically inspired, they are mainly designed as engineering solutions to specific problems in NLP. On the other hand, little work was done to integrate neural models of language into comprehensive cognitive models compatible with current knowledge of how verbal information is stored and processed in the brain, i.e. with verbal working memory models. Miikkulainen [13,19] and Fidelman et al. [20] presented a cognitive neural architecture able to parse script-based stories, to store them in episodic memory, to generate paraphrases of the narratives, and to answers questions about them. Their model was tested on a small corpus of nine scripts, each of which consisted of 4-7 sentences.

Dominey and Hinaut [21,22] proposed a neural model of brain areas involved in language processing, able to learn grammatical constructions and to generalize the acquired knowledge to novel constructions. In their work, language understanding is identified as the ability to recognize the thematic role of the open-class words in the surface form of sentences, and meaning is interpreted as a mapping from the surface form to a functional form of sentences. This notion of understanding is not sufficient for the purpose of the present work, which is more focused on the elaboration of verbal information in the working memory and on the procedural knowledge involved in question answering and, more generally, in communicative interactions.

## 1.1  Working memory models

Although there are different perspectives regarding the organization of memory in the human brain, all approaches recognize at least two types of memory: the *short-term memory* (STM) and the *long-term memory* (LTM). STM can be defined as the capacity of the human mind to hold a limited amount of information in a readily accessible state for a short period of time. In contrast, LTM is a large repository of knowledge and of information on prior events, which can be stored in the mind for long periods of time. The term *working-memory* (WM) has been defined in different ways, however most researchers assume that WM includes (at least) the STM and the processing mechanisms used for temporarily storing and manipulating information in the STM.

In 1974, Baddeley and Hitch [23] proposed a working memory model composed of three main components: a *central executive* and two slave systems, i.e. the *phonological loop* and the *visuo-spatial sketchpad*. The central executive operates as a supervisory system by controlling the flow of information from and to the slave systems. The slave systems are responsible for short-term maintenance of information: the phonological loop stores verbal content, while the visuo-spatial sketchpad stores visual and spatial information.

In 2000, Baddeley [24] extended this model by adding a third slave system, the *episodic buffer*, which binds information from different domains (phonological, visual, spatial, semantic) to form integrated units of information with chronological ordering. Fig 1 shows a schematic diagram of this model.

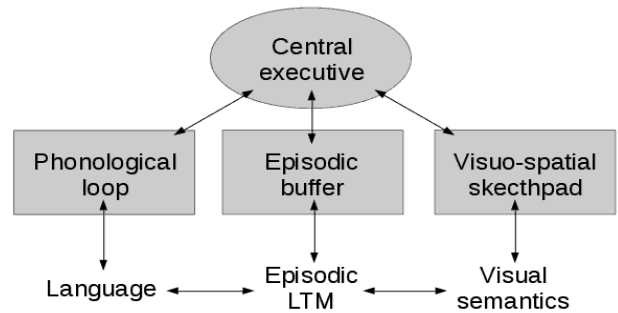

**Fig 1.** The Baddeley's model of the working memory.

The Baddeley's model is supported by evidences from experimental psychology, neuropsychology and cognitive neuroscience (see Ref. [25] for a review). However, some criticism have been raised and alternative models have been proposed. Cowan [26] proposed a working memory model in which the LTM was not a separate component, but a part of the working memory. Cowan's model consists of four components: a central executive, a LTM, an *activated memory* and a *focus of attention*. The central executive directs attention and controls voluntary processing. The activated memory is the subset of LTM in a state of temporal activation, and it can hold a large number of activated elements. The focus of attention is a subset of the activated memory. It has a limited capacity and can hold up to about four independent items or chunks. According to Baddeley, the differences between his view and that of Cowan are mainly in "emphasis and terminology" [25]. In particular, the episodic buffer of his model has a similar role to Cowan's focus of attention. McElree [27] suggested a focus of attention limited to a single chunk. Oberauer [28] proposed a model that distinguishes three states of representations in WM: the activated part of LTM, the *region of direct access* and the focus of attention. The region of direct access roughly corresponds to the broader focus of attention in Cowan's model, with a scope of about four chunks. The focus of attention in Oberauer's model corresponds to the single-chunk focus of McElree's model. The function of the focus of attention is to select a single item or chunk from the direct-access region.

## 1.2  The mental action sequence

In classical tasks used to study working memory capacity [28], a subject is asked to hold in mind a short sequence of digits and to perform some simple process on each of these digits (or on a subset), for example adding the number two to each digit. Consider, for instance, the following task:

"*Add the number two to each of the following digits: 6 3 9 4*"

We assume that the subject has memorized additions with small numbers in LTM, so that the cognitive load for a single addition is small. The sequence of mental operations that are performed by the subject can be the following:

1) transfer the 4 digits 6 3 9 4 to the phonological store;

2) transfer the first digit (6) to the focus of attention;

3) use this digit as a cue to retrieve the appropriate operation from the LTM; for instance, the following sentence can be retrieved from the LTM and transferred to the phonological store: "*six plus two equal eight*"; clearly, this is just an example, and the same result could be retrieved from LTM in other ways;

4) transfer the result ("eight") to the focus of attention, and use it for speech production;

5) transfer again the four digits 6 3 9 4 to the phonological store;

6) transfer the second digit (3) to the focus of attention

.....

and so on, until the last digit is processed. Additionally, several studies [29,30] suggest that the task goal should be stored in the working memory in some directly accessible form. Therefore, the previous sequence should be extended by including at the beginning, before step 1, two other operations, such as:

a) transfer the phrase "*add the number two*" to the phonological store;

b) transfer this phrase (or some coded form of it) to a goal-task store.

In the next section we will illustrate how the "mental action sequence" (a,b,1-6) is implemented in our model. In the following sections, we will also demonstrate that a broad range of tasks in human language processing can be performed using iterations of this basic action sequence. A minimal system that can perform this sequence should include (at least) the following components:

- a phonological store;

- a focus of attention;

- a retrieval structure that uses the focus of attention as a cue to retrieve information from LTM;

- a goal store (i.e. a goal stack in our model, as in many other cognitive architectures);

- a supervising system that controls the flow of information among the other components, i.e. a central executive.

At this point, one may wonder why a neural architecture is necessary to model this process. Apart from the obvious consideration that our brain is a neural architecture, why a symbolic model is not enough? What we try to emphasize in our work is that the decision processes operated by the central executive are not rule-based process, they are statistical decision processes. In our model, the central executive is a neural network that takes as input the signal from the STM components (the internal state) and provides as output mental actions that direct the flow of information among the slave systems. Therefore, the central executive should comprise a state-action association system. If the central executive was not a statistical tool, the system would not be able to generalize. But how might the generalization arise in the previous example? Suppose that an artificial model of the working memory was trained to respond to the "add the number two" task described above, and that it is tested on a similar task, but with different numbers:

"*add the number three to each of the following digits: 7 8 2 5*"

Since this sentence is similar to that of the first task, the central executive will provide the same output, i.e. the same mental-action sequence.

Through this sequence, the system will extract the phrase "add the number three" and push it in the goal stack, then it will transfer the sequence "7 8 2 5" to the phonological store, it will transfer the first number ("7") to the focus of attention and use it as a cue to retrieve information from LTM. Now we come to another question: why the retrieval process should be modeled using a neural architecture, or more generally why the retrieval process should be described as a statistical process? In principle there could be thousands of phrases that could be retrieved from LTM using the digit "7" as a cue. How can the system choose the appropriate phrase among them? The system can recognize that some of the phrases that can be retrieved from LTM using the digit "7" as a cue are similar to the one retrieved during the training stage, which was:

"*six plus two equal eight*" (phrase 1)

For instance

"*seven plus two equal nine*" (phrase 2), or

"*seven plus three equal ten*" (phrase 3)

contain the cue ("7") and are similar to phrase 1, in the sense that both phrase 2 and phrase 3 are close to phrase 1 in the input space of the state-action association system. Unfortunately, phrase 2 is closer. If the choice was based solely on similarity with the phrase retrieved during training, the system would choose phrase 2, and following the same action sequence of the training example, it would give a wrong answer, i.e. "nine" instead of "ten". In our model, the generalization capabilities are supported by a "comparison structure", which is an additional component of the STM that recognizes similarities among elements of different STM components. For instance, it can recognize that one word in the phonological store is equal to a word of the phrase stored in the goal stack. In our example, the comparison structure allows the system to recognize that the third word of phrase 2 ("three") is equal to the fourth word in the goal phrase "add the number three". In a simple neural model of the comparison structure, the neurons that compare those two words will be activated. Our model includes a comparison structure, which is part of the input to the state-action association system of the central executive. We will show that the connections from the comparison structure to the central executive are weighted more than the connections from the phonological store to the central executive, therefore in the above example the system will select phrase 3 rather than phrase 2, and it will give the correct answer.

## 1.3 Localization of the verbal working memory in the brain

Localization of brain areas that are involved in language comprehension and production requires the combination of findings from neuroimaging and psycholinguistic research. Several studies on the functional neuroanatomy of language indicate that both semantic and syntactic processes involve mainly the left frontal cortex and part of the temporal cortex. The left frontal cortex is considered to be responsible for strategic and executive aspects of language processing. The left temporal cortex supports the processes that identify phonetic and lexical elements. It is involved in storage and retrieval of phonological, syntactic and semantic information form memory.

All classical neurobiological models of language attribute a fundamental role to the Broca's area, which includes Brodmann's areas (BA) 44 and 45, in the left frontal cortex. Several studies show that BA 47 and the ventral part of BA 6 are also involved in language processing tasks [31-33]. The language-relevant part of the frontal cortex is thus the left inferior frontal gyrus (LIFG) which comprises BA 44, 45, 47 and 6. Results from neuroimaging and psycholinguistic studies show that LIFG is involved in the unification operations required for binding individual words into larger structures [34,35]. Hagoort [34] proposes a model that distinguishes three functional components of language processing: memory, unification and control. Fig 2 shows the main areas of the cortex that support the three components.

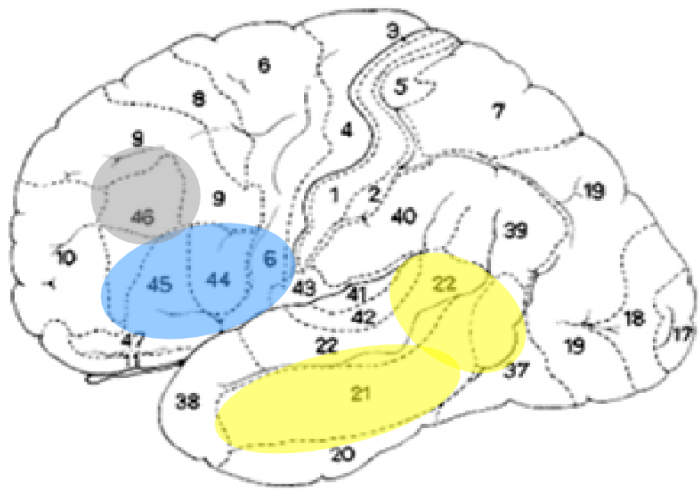

**Fig 2.** Localization of the Memory (yellow), Unification (cyan) and Control (grey) components of the MUC model proposed by Hagoort. The Brodmann's areas are marked by numbers.

## 2 Methods

### 2.1 The ANNABELL model

The model presented in this work, called ANNABELL (Artificial Neural Network with Adaptive Behavior Exploited for Language Learning), is a cognitive neural architecture, designed to help understand the cognitive processes involved in early language development. The source code of the software, the User Guide and the datasets used for its validation are available in the ANNABELL web site at https://github.com/golosio/annabell/wiki .

The global organization of the system is compatible with the multicomponent working memory (M-WM) framework proposed by Baddeley. However, our work is focused on the role of executive functions in language processing tasks, and not on many other important questions concerning WM, as those related to working memory capacity or information maintenance in STM. Therefore, for the sake of simplicity, our model does not take into account many effects that are of central importance for working memory theories, as for instance phonological/semantic similarity, word length effect, recency, and other effects in serial and free recall tasks. We also do not take a position in the controversy on whether information in the phonological store is maintained by passive storage or by active rehearsal, and it is again for reasons of simplicity that we have chosen passive maintenance.

The building blocks of the model are artificial neurons. The system is based on the concept of sparse-signal map (SSM). A SSM is simply an ANN that has only a small fraction of all neurons active at a given time. The advantage of this representation is that it can be implemented in a very efficient way both in terms of computation time and in terms of memory usage, therefore it can partially compensate for the relatively limited parallelism of available hardware compared to the biological brain. The design of the neuron model focused on computational efficiency rather than biological details. It is important to point out that the purpose of this approach it not an engineering solution to the human-machine dialogue problem, but a cognitive model of how verbal information is processed in the brain. Computational efficiency is necessary for building a large-scale neural model of the verbal working memory, able to sustain a long training procedure on a relatively large database.

The system is composed by several SSMs, connected to each other either by fixed-weight or by variable-weight (learnable) connections. The latter ones are updated through a discrete version of the Hebbian learning rule. Most of the learnable connections are virtual: they are not actually allocated in memory, unless their default weight value is modified by the Hebbian mechanism. As the signal is sparse, only a small fraction of the neurons is active at a given time, therefore most learnable connections remain virtual. With this approach memory requirements and, most importantly, computational time are greatly reduced compared to conventional techniques. The proposed system is faster by at least three orders of magnitude compared to other large-scale neural systems.

The communication between the system and the human interlocutor is achieved through an interface that converts words into input patterns, submits them one by one to the system, extracts output patterns and converts them to words. The network architecture is designed in such a way that the system can process phrases using *mental actions*, which are elementary operations on word groups and phrases that are used, for instance, for acquiring the words of the input phrases, for memorizing phrases, for extracting word groups from the working phrase, for retrieving memorized phrases from word groups through an association mechanism, etc. Such actions are performed by special neurons, called *mental action neurons*, which can control the flow of signal between different subnetworks. A key feature of the model is that the connections that are affected by the reward mechanism are connected to mental action neurons, rather than being directly connected to output words or phrases. In this way, the system learns preferentially to build the output through sequences of elementary operations on word groups or phrases. This type of architecture underpins the generalization capabilities of the system.

The system was implemented on a PC equipped with a high-performance GPU (graphics processing unit) NVIDIA Kepler GK104 having 1536 processing units (called cores). GPUs are programmable logic chips that are widely used not only for graphical applications, but more generally for high-performance-computing applications that require a high degree of parallelism. The current version of the system is composed by 2.1 million neurons, interconnected through 33 billion virtual connections. At the end of the complete learning process described in this work, the number of real (allocated) connections was 27 million. The size of the system is comparable to that of the neural architecture described in Ref. [8], although our model privileges computational efficiency over biological details. The ability to perform real time communication and the large scale of the network make our system adequate for sustaining a relatively long developmental process (this property is called open-ended, cumulative learning in developmental robotics [36] ). The system is being trained through an approach that, compared to those used for other artificial systems, is much more similar to children language training. This process is conducted by personifying the system as a child in a virtual social environment. The validation of its performance is inspired by the literature on early language assessment. Test sessions are used to assess syntax, semantics, pragmatic language skills, communicative interactions, language processing skills and comprehension of sentence structure.

## 2.2 Learning mechanisms and signal flow control

The ANNABELL system is entirely composed of interconnected artificial neurons, and all processes are achieved at the neural level. Although different subsystems can be distinguished by their function, the whole system has a unitary structure. The subnetworks are arranged in layers that determine the update order, with both forward and backward (recurrent) connections among different layers.

The system uses a standard artificial neuron model. The neurons are connected among each other by directional weighted connections (links). Three types of connections are used:

- fixed-weight connections, which do not change during the learning process;
- variable-weight (learnable) connections, which are modified by the learning process;
- forcing connections, which are variable-weight connections that have a positive or negative weight much higher in absolute value than that of the other two connection types, thus they can force the target neurons to a high-level or to a low-level state.

The total input signal of each neuron is evaluated as the weighted sum of the signals coming from its input connections:

$$y_i = \sum_{j \in S_i} w_{ij} o_j + b_i$$

where $i$ is the neuron index, $y_i$ is its total input signal, $S_i$ is the set of neurons that are connected to the other ends of its input connections, $j$ is an index that runs on the set $S_i$, $w_{ij}$ are the weights of the input connections, $o_j$ are the output signals of the neurons connected to its input, and $b_i$ is a bias signal. The neuron output is computed from the total input by a nonlinear activation function [37]:

$$o_i = f(y_i)$$

which approaches zero as $y_i$ tends to minus infinity, or one as $y_i$ tends to plus infinity. Two types of activation functions are used in the model, i.e. the Heaviside step function for the neurons that receive their input from fixed-weight connections, and the logistic function [37] for the neurons that receive it from variable-weight connections.

In the subnetworks that have learnable input connections, the inhibitory competition among neurons is modeled using the *k-winner-take-all* rule, i.e. the *k* neurons with the highest activation state are switched on, while all the remaining neurons are left off. This rule provides a computationally effective approximation of the activation dynamics produced by inhibitory interneurons [38]. The Hebbian theory provides a theoretical basis for the learning mechanisms in biological neural networks [39,37]. According to this theory, the strength of the synaptic junction between two neurons is increased when the outputs of the two neurons are strongly correlated, i.e. when the two neurons *fire together*. In our model, the learnable connections are modified through a discrete version of the Hebbian learning rule (DHL rule), combined with the k-winner-take-all rule: the connection weight is modified only if the postsynaptic neuron is one of the *k* winners of the k-winner-take-all competition; if the presynaptic neuron at the other end of the connection is in the same activation state as the winner neuron (i.e. in the high level state "on") the connection weight is saturated to its maximum value. In the opposite case, it is saturated to its minimum value. A detailed description of the learning algorithms and of the statistical properties of the state-action association system is provided in Appendix D.

The elementary operation of the system on word groups, phrase buffers and other SSMs, called *mental actions,* are triggered by special neurons that will be called *action neurons*. Each action neuron can activate one or more *gatekeeper* neurons, which control the flow of signal between different subnetworks. Neural gating mechanisms play an important role in the cortex and in other regions of the brain [40]. They rely on the action of bistable neurons, i.e. neurons that can oscillate between a quiescent "down" state, associated with a hyperpolarized membrane potential, and an "up state", characterized by a membrane potential that is just below the cell's firing threshold. The gatekeeper neurons can modulate the membrane potential of the bistable neurons, shifting them from the "down" state to the "up" state and vice versa. Different types of neural gating mechanisms have been observed in the brain. Fig 3 represents the type of gating mechanism that is exploited in our model. In this example, a gatekeeper neuron is fully connected to a set of bistable neurons. When the gating signal is "off", the gate is closed: the bistable neurons are in the "down" state, and they do not respond to the input signal. Conversely, when the gating signal is "on" the gate is open: the bistable neurons are in the "up" state and they transmit the input signal to the second set of neurons. The bistable neurons therefore perform a type of biological AND relative to their inputs.

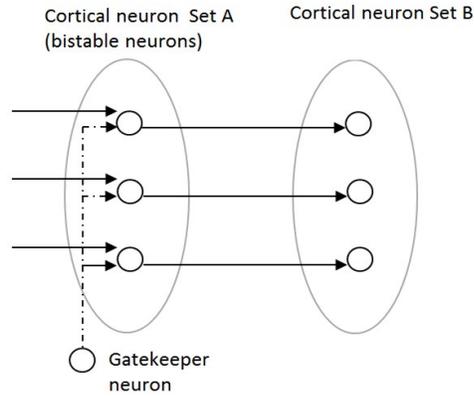

**Fig 3.** Type of neural gating mechanism used in the ANNABELL model. (adapted from Ref. [40]).

In the ANNABELL model, the gatekeeper neurons are generally fully connected to one or more SSMs, and they can enhance or suppress the activation state of the SSM neurons, acting in a similar way to a change in the bias signal. In this manner they can control the flow of signal from one part of the system to another. The mental action neurons and the gatekeeper neurons are based on the same simple neuron model used for all neurons of the system. Their specialization is only a result of the way how they are connected to other subnetworks. In particular, the action neurons receive the input signal from the state-action association subnetwork, and they send their output signal to the gatekeeper neurons.

The input and the output connections of the state-action association network follow a distributed model, i.e. the state-action association network is fully connected to the subnetworks that represent the internal state of the system (input) and to the action neurons (output). The input and output connections of the state-action association SSM are updated through the DHL rule. On the other hand, the connections between the action neurons and the gatekeeper neurons have fixed, predetermined weights, in such a way that each action neuron corresponds to a well-defined operation. The output connections of the gatekeeper neurons are generally fully connected to one or more subnetworks, in such a way that they can allow or inhibit the flow of signal through such subnetworks. A key feature of the ANNABELL system that is particularly important for his generalization capabilities is that the learnable connections that are affected by the reward (i.e. the connections of the state-action association SSM) are connected to action neurons, rather than being directly connected to output words or phrases. In this way, the system learns preferentially to build the output through sequences of elementary operations on word groups or phrases.

## 2.3 Global organization of the model

The global organization of our model is compatible with the M-WM framework. This section presents an overview of the system architecture and operating modes. The Supporting Information document *The ANNABELL system architecture* (available at http://www.neuralsystems.net/annabell/files) provides a detailed description of the architecture, while Appendix C describes in details how the neural activation patterns evolve and how the connection weights are modified on a concrete example. However we must point out that the details of the implementation and further divisions in subcomponents, as described in Appendix C and in *The ANNABELL system architecture*, mainly respond to the need of building a neural-network model suitable for simulations that produce cognitively relevant behavior, and should not be considered as a premature attempt to map the model architecture to neural circuits in the biological brain.

The ANNABELL model comprises four main components, as shown in Fig 4: a verbal short-term memory (STM), a verbal long-term memory (LTM), a central executive (CE) and a reward structure.

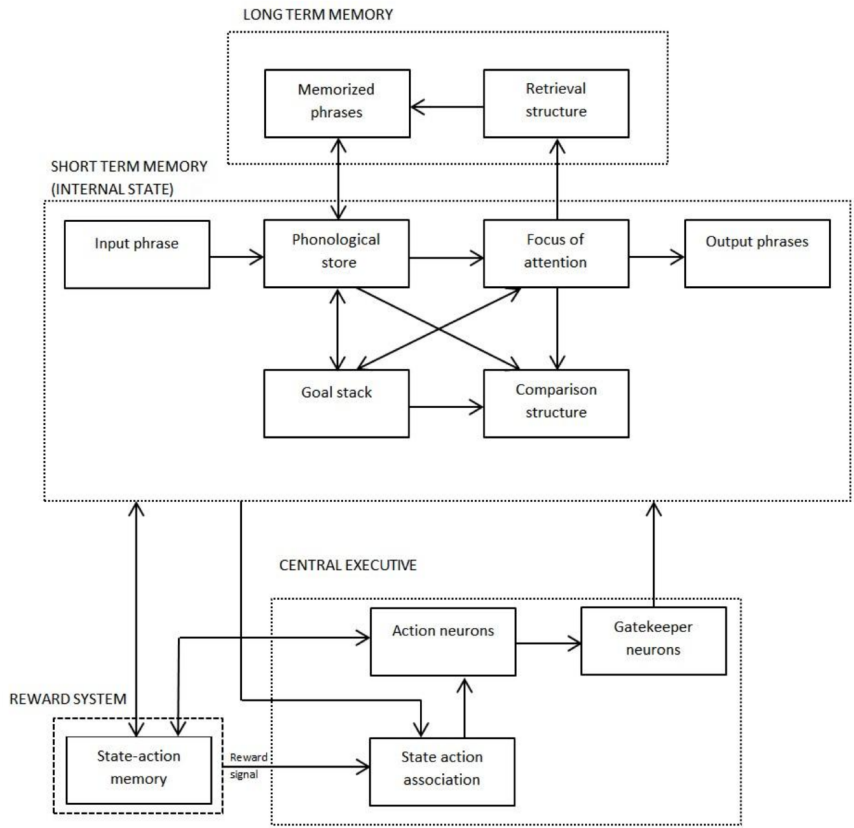

**Fig 4.** Schematic diagram of the ANNABELL system main components.

The STM includes a phonological store, a focus of attention, a goal stack and a comparison structure. The phonological store maintains the working phrase. The focus of attention holds up to about four words. It is involved in several functions, including language production planning, and it is also used as a cue for retrieving information from LTM. For reasons of simplicity, our model does not include a visuo-spatial system or other types of sensory inputs; therefore, unlike Baddeley's episodic buffer, the focus of attention of our model can hold only verbal content. The goal stack is a structure for storing goal chunks that contribute to decision-making processes. The comparison structure recognizes similarities among words in the phonological store, in the focus of attention and in the goal stack, and is also used for decision-making processes. The LTM includes a structure for memorizing phrases and a retrieval structure that uses the focus of attention as a cue for retrieving memorized phrases. The CE is a supervisory system that controls all decision-dependent processes through neural gating mechanisms.

It includes a state-action association system, a set of action neurons and a set of gatekeeper neurons. The state-action association system is a structure that is trained by a rewarding procedure to associate mental actions to the internal states of the system. The gatekeeper neurons are neurons that can control the flow of signal between different subnetworks by acting in a similar way as an increase or a decrease of the bias signal. The mental-action neurons are used to perform elementary operations on phrases, as increasing the phrase index, extracting a single word from the working-phrase buffer and mapping it to the word-group buffer, retrieving a memorized phrase from a word group, storing the working phrase in the goal stack, etc. Each (mental) action neuron corresponds to a well-defined action, which can be executed by activating the state of one or more gatekeeper neurons. The system can perform three types of actions.

1. Acquisition actions. Those actions are used during the acquisition and during the association phases, for acquiring the input phrases, memorizing them and building the associations between word groups and memorized phrases.
2. Elaboration actions. Those actions are used during the exploration and during the exploitation phases, for extracting word groups from the working phrase, for retrieving memorized phrases from word groups through the association mechanism, for retrieving memorized phrases belonging to the same context, for composing output phrases.
3. Reward actions. Those actions are used by the rewarding system and can be executed in parallel to the elaboration actions. They are used for memorizing the state-action sequences produced during the exploration and during the exploitation phases, for retrieving such sequences after a reward signal and for triggering the changes of the state-action-association connection weights.

A complete list of the actions is presented in *The ANNABELL system architecture*.

The reward structure memorizes and retrieves the sequences of internal states of the system and the mental actions performed by the system (state-action sequences). When an exploration phase produces a target output, the reward structure retrieves the state-action sequence, and it rewards the association between each internal state and the corresponding mental action, by triggering synaptic changes of the state-action association output connections.

The ANNABELL system is composed of several subnetworks. Fig 5 represents a schematic diagram of the main subnetworks in the STM and in the LTM. Each rectangular block in this diagram represents a subnetwork composed by interconnected artificial neurons. A detailed description of the system architecture is provided in *The ANNABELL system architecture*.

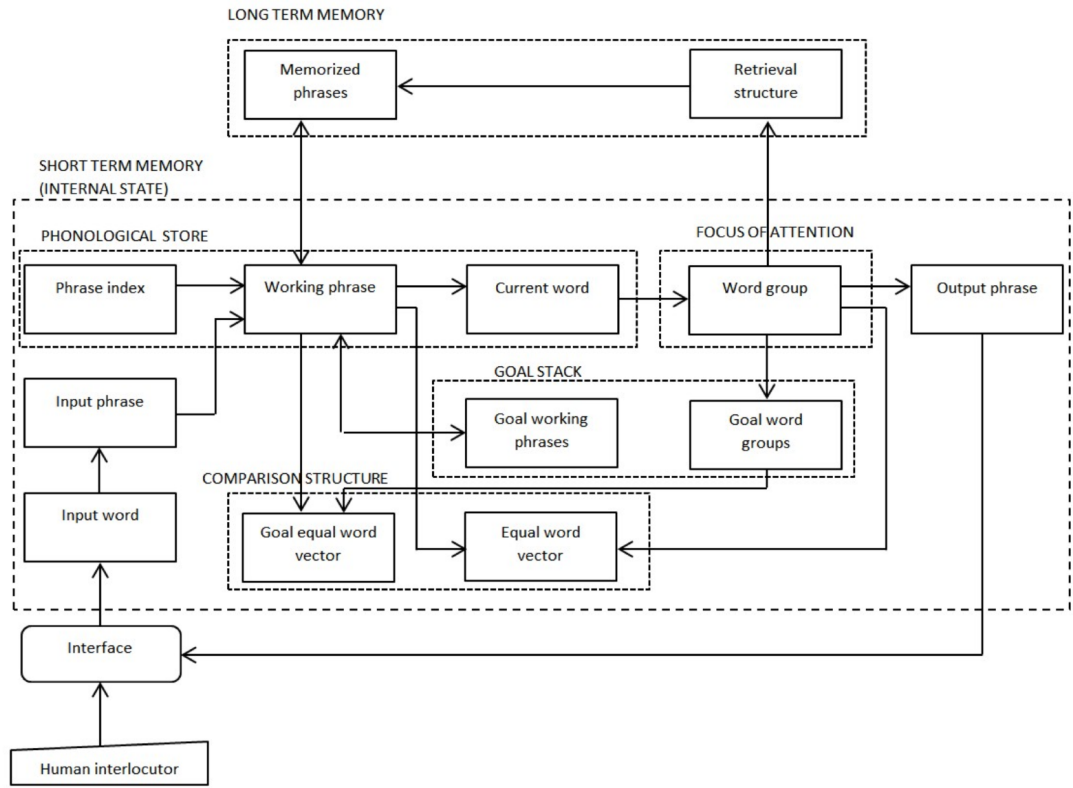

**Fig 5** Schematic diagram of the main system architecture. Each rectangle represent a subnetwork, which is composed by interconnected artificial neurons. Only the main subnetworks are represented in this diagram. The arrows that join the rectangles represent directional connections among neurons of different subnetworks.

The communication is achieved through an user interface between the human interlocutor and the system. The interface converts words into input patterns and submits them one by one to the system, extracts output patterns and convert them to words. It also sends reward signals to the system when prompted by the human. The interface includes a monitor tool that can be used to display the content of the SSMs that compose the system.

The system can work in five operating modes, which are briefly described below.
1. **Acquisition.** In this operating mode, the words of a phrase are acquired one by one and stored in the input-phrase buffer.

2. **Association.** In this operating mode, the input phrase is copied from the input-phrase buffer to the working-phrase buffer and it is stored in a long-term memory (represented by the block *Memorized Phrases* in Fig 5). After that, all possible groups of contiguous words (with maximum four words in a group) are extracted from the working phrase and copied to the word-group buffer, and the association between the word group and the whole phrase is memorized in a long term memory (the block *retrieval structure* in Fig 5).
3. **Exploration.** In this mode the system executes partially random sequences of elementary operations (mental actions) on word groups and phrase buffers. The basic action sequence used during the exploration operating mode is the following:
    - *W_FROM_WK:* initializes the phrase index (PhI) to zero, to prepare the extraction of words from the working-phrase buffer;
    - *NEXT_W* ($N_1$ times): skips $N_1$ words of the working phrase buffer;
    - *FLUSH_WG*: clears the content of the word-group buffer;
    - *GET_W, NEXT_W* ($N_2$ times): copies $N_2$ consecutive words from the working phrase buffer to the word-group buffer;
    - *WG_OUT* (0/1 times): copies the word-group buffer content to the output buffer;
    - *RETR_AS* (0/1 times): retrieves a phrase associated to the word group by the association mechanism.

    $N_1$ and $N_2$ are random integer numbers. $N_1$ can eventually be null, while $N_2$ must be greater than or equal to one. The range of $N_1$ and $N_2$ depends on the maximum phrase size (ten words in the current implementation). Additionally, the system can eventually execute the following actions:
    - *GET_START_PH* (0/1 times): retrieves the starting phrase in the same context of the working phrase;
    - *GET_NEXT_PH* ($N_3$ times): retrieves sequentially phrases belonging to the same context;

    The human interlocutor can suggest to the system a target phrase or a target word group. The basic action sequence can be iterated more times, until the system produces an output. If the output does not correspond to the target output, the whole process is restarted. The exploration is terminated when it produces the target phrase / target word group, or when the number of iterations becomes greater than a predefined limit.

When the working phrase indicates a task that cannot be executed immediately, it can be set as a goal by inserting it in a SSM that acts as a *goal stack* with the action *PUSH_GOAL*. When the goal is reached, the phrase can be removed from the stack with the action *DROP_GOAL*.

4. **Reward.** When the exploration process produces a phrase or a word group that the teacher recognizes as worth to be rewarded (target phrase or target word group) he can activate a rewarding procedure. In this operating mode the system retrieves the state-action sequence that led to the target phrase / target word group. The association between each state of the sequence and each corresponding action is rewarded by changing the connection weights of the state-action association SSM through the DHL rule.

5. **Exploitation.** In this operating mode the state-action association SSM, trained by the rewarding procedure, is used to associate a mental action to each system state. The state-action-association SSM is updated through the k-winner-take-all rule. It receives as input the internal state of the system (represented by a dashed rectangle in Fig 5), and it sends its output to the elaboration-actions SSM, which is updated through the (one) winner-take-all rule. In this way a single elaboration action is selected, the one that is more represented among the outputs of the k winners of the state-action-association SSM.

Appendix C describes in details, on a concrete example, how the neural activation patterns evolve, how the connection weights are modified during training, and how these weight changes make the system able to generalize the acquired knowledge to new sentences.

## 2.4 The database

The database of sentences used for training and testing the system is organized in five datasets, each devoted to a thematic group, i.e. people, parts of the body, categorization, communicative interactions and movement in a text-based virtual environment. Each of those datasets includes declarative sentences, conversational sentences and interrogative sentences. Declarative sentences are used to give some information to the system without expecting a response. As the system has no sensory input, apart from that provided by the text-based interface, all the information must be provided in the form of input sentences. Interrogative sentences are questions that expect an answer from the system. In the training stage, for each question the teacher suggests the associations that can be used to build a valid answer. In the test stages, the questions are used to verify whether the system is able to generalize what it learned during the training phase. An answer is considered to be correct only if it is both syntactically and semantically correct.

Conversational sentences that expect a turn taking from the system are treated in the same way as the questions: for this type of sentences, in the training stage the teacher suggests response sentences that are appropriate for the conversation. On the other hand, conversational sentences that do not expect a turn taking are treated as declarative sentences.

### 2.4.1 The *people* dataset

The first dataset is devoted to the subject *people*, and it is partially inspired by the Language Development Survey work of Rescorla et al. [41,42]. The sentences of this dataset have been prepared by personifying the system in a four years old little girl in her social environment, which includes the two parents, a sister, a friend, two cousins, the four grandparents, two aunts, two uncles and six other children, for a total number of twenty persons. Four of those persons, namely the two parents, the sister and the friend, are considered to have a closer relationship to the system, which means that the dataset provides more information for those four persons than for the others. In some cases, the two cousins are also included in the group of closer persons. Some sentences depend on the possible relationships between the persons and the system. In such case, we distinguish nine types of relationships, i.e. father, mother, sister, friend, cousin, grandmother, grandfather, aunt and uncle. The six other children are included in the social environment mainly for training and evaluating the system in age comparison tasks. Some declarative sentences (*how-to* sentences) are used to provide prescriptions on how to accomplish some specific tasks, as for instance

*to answer if someone is younger or older than you, you should compare your age with his age*

or to express language rules in a simple verbal form, as

*the possessive pronoun for a woman is her* .

Table 1 shows the types of declarative sentences used in the people dataset. The total number of declarative sentences in this dataset is 225.

The questions used in the people dataset are also inspired by the work of Rescorla [41,42], and they are appropriate for a preschool child, as in the following examples:

*what does your father do?*

*what games do you like?*

*do you have a sister?*

*is Dad older than Mum?*

etc. A full list of the declarative sentences and of the questions can be found in the files that are distributed with the software package. They explore the meaning of words, but they are also used to train the system for language and reasoning skills, as:

- use of personal and possessive pronouns;
- answer to polar (yes/no) questions, alternative (choice) questions, wh-questions and question-like imperative sentences (e.g. tell me);
- counting and comparing numbers, as for instance in age comparison:

    *is Letizia older or younger than your sister?*

- learning language rules:

    *the possessive pronoun for a female person is her*

**Table 1.** Sentences of the *people* dataset. The social environment described in this dataset includes twenty persons. In the second column, <person> can be "Mum", "Dad", or the name of one of the other eighteen persons, <relationship> can be "father", "mother", "sister", "friend", "cousin", "Grandma", "Grandpa", "aunt" or "uncle". <number> can be a number or, in row 21, also "some" or "many". The "(s)" denotes the possibility of a plural form. In row 15, <verb> and <complement> describe the profession in terms understandable for a preschool child, e.g. "the journalist writes in the newspaper". The sentences in row 24 use the present progressive, as in "Susan is reading a book". The sentences in row 25 (*how-to* sentences) are verbal prescriptions, expressed through the natural language, that are used to instruct the system on how to perform specific tasks in language processing.

| | **Sentence structure** | **Parents** | **Sister** | **Friend** | **Cousins** | **Grand-parents** | **Aunts/uncles** | **Other children** | **Total** |
|---|---|---|---|---|---|---|---|---|---|
| 1 | <person> is your <relationship> | 2 | 1 | 1 | 2 | 4 | 4 | 0 | 14 |
| 2 | You have <number> <relationship>(s) | 2 | 1 | 1 | 1 | 2 | 2 | 0 | 9 |
| 3 | <person> is a woman/man/girl/boy | 2 | 1 | 1 | 2 | 4 | 4 | 6 | 20 |
| 4 | <person> is a <profession> | 2 | 0 | 0 | 0 | 4 | 4 | 0 | 10 |
| 5 | <person> goes to the school/kindergarten | 0 | 1 | 1 | 2 | 0 | 0 | 3 | 7 |
| 6 | <person> has a <noun> | 2 | 1 | 1 | 0 | 0 | 0 | 0 | 4 |
| 7 | <person> does not have a <noun> | 2 | 1 | 1 | 0 | 0 | 0 | 0 | 4 |
| 8 | Your <relationship>'s name is <name> | 2 | 1 | 1 | 0 | 0 | 0 | 0 | 4 |
| 9 | <person> likes <noun>(s) | 4 | 2 | 2 | 0 | 0 | 0 | 0 | 8 |
| 10 | <person> likes to <verb> … | 4 | 2 | 2 | 0 | 0 | 0 | 0 | 8 |
| 11 | <person> does not like <noun>(s) | 4 | 2 | 2 | 0 | 0 | 0 | 0 | 8 |
| 12 | <person> does not like to <verb> … | 4 | 2 | 2 | 0 | 0 | 0 | 0 | 8 |
| 13 | <person> is <number> years old | 2 | 1 | 1 | 2 | 4 | 4 | 6 | 20 |
| 14 | <person>'s favorite <noun> is … | 4 | 2 | 2 | 4 | 0 | 0 | 0 | 12 |
| 15 | The <profession> <verb> <complement> | - | - | - | - | - | - | - | 8 |
| 16 | You like <noun>(s) | - | - | - | - | - | - | - | 6 |
| 17 | You like to <verb> … | - | - | - | - | - | - | - | 6 |
| 18 | You do not like <noun>(s) | - | - | - | - | - | - | - | 6 |
| 19 | You do not like to <verb> … | - | - | - | - | - | - | - | 6 |
| 20 | Your favorite <noun> is … | - | - | - | - | - | - | - | 4 |
| 21 | You have <number> <noun>(s) | - | - | - | - | - | - | - | 4 |
| 22 | You do not have a <noun> | - | - | - | - | - | - | - | 4 |
| 23 | Women/men/girls/boys like to … | - | - | - | - | - | - | - | 7 |
| 24 | <person> is <verb>-ing <complement> (present progressive) | 2 | 1 | 1 | 2 | - | - | - | 6 |
| 25 | *How-to* sentences | - | - | - | - | - | - | - | 28 |
| 26 | Other sentences | - | - | - | - | - | - | - | 4 |
| | | | | | | | | **Total** | 225 |

The following question/answer example illustrates some of the abilities acquired by the system:

Q: *is your friend younger than you?*

A: *no, she is older.*

The system is able to answer to the question Q by following a line of reasoning that it has learned through the communication with the human, thanks to its adaptive behavior. The system uses the past experience listed below.

1) The system has been taught to count;

2) The system has been taught to decide whether another child is younger or older than the girl that it impersonates, through the following phrases:

*to answer if someone is younger or older than you, you should compare your age with his age*

3) The system has learned the age of the girl that it impersonates:

*you are four years old*

4) The system has learned that the words "your friend" refer to the friend Letizia

*Letizia is your friend*

5) The system knows the age of Letizia:

*Letizia is five years old*

6) The system has learned how to use personal pronouns, therefore it can answer using the personal pronoun *she* instead of the name *Letizia*.

The teacher taught the system to answer questions similar to the question Q, guiding it through a series of mental operations (associations and extractions of word groups from sentences), through the exploration-reward method described previously. At this point the system is able to generalize the procedure and to answer questions similar to those used for training.

It is important to emphasize that this whole process takes place in the system at a subsymbolic (neural) level and that phrase memorization and learning take place in the form of changes of the weights of the connections between neurons through the DHL rule. The examples shown in Appendix A show in more detail how the system is trained to answer to a question.

### 2.4.2 The *parts of the body* dataset

The second dataset is devoted to the main parts of the body, and it is also partially based on the words of this subject category included in the Language Development Survey. Through this dataset, the system is trained to recognize the definition of a word as well as different ways to specify the location of an object. After the training, the system should be able to answer questions of the type *what is* and *where is*. Table 2 represents the type of declarative sentences used in this dataset. Thirty-three body parts are considered. For each of them, a declarative sentence provides a simple definition in a form that should be understandable for a preschool child. Other sentences specify the locations of the body parts. It can be observed that in this case the correspondence between body parts and sentences is not one-to-one, because the location of a body part can be described in more than one way. Eight declarative sentences describe in simple terms what is the function of some body parts, e.g.

*with your legs you can walk, run and jump*

and finally, six sentences are *how-to* sentences. The total number of declarative sentences in the body parts datasets is 122.

**Table 2.** Sentences of the *parts-of-the-body* dataset. Thirty-three body parts are included in the dataset. In the first column, <part> is the name of a body part. The "(s)" refers to a possible plural form.

| Body part definitions – singular form | Number of sentences |
|---|---|
| the <part> is the part of ... | 13 |
| the <part> is the joint ... | 4 |
| the <part> is the <adjective> part of ... | 5 |
| the <part> is the ...(excluding previous forms) | 9 |
| Other cases | 2 |
| Subtotal: | 33 |
| **Body part definitions – plural form** | |
| the <part>s are the parts of ... | 8 |
| the <part>s are the joints ... | 4 |
| the <part>s are the ...(excluding previous forms) | 7 |
| Subtotal: | 19 |
| **Body part locations – singular form** | |
| the <part> is on the ... | 3 |
| the <part> is at the ... | 4 |
| the <part> is in the ... | 9 |
| the <part> is between the ... | 9 |
| the <part> is attached to ... | 5 |
| the <part> is above the ... | 1 |
| the <part> is inside the ... | 4 |
| Other cases | 2 |
| Subtotal: | 37 |
| **Body part locations – plural form** | |
| the <part>s are between the ... | 1 |
| the <part>s are attached to ... | 5 |
| the <part>s are in the ... | 1 |
| the <part>s are above the ... | 1 |
| the <part>s are at the ... | 4 |
| the <part>s are on the ... | 1 |
| Other forms | 2 |
| Subtotal: | 15 |
| **Other sentences** | |
| the <part> connects the ... | 4 |
| with your <part>(s) you can ... | 8 |
| *How-to* sentences | 6 |
| Subtotal: | 12 |
| **Total:** | **122** |

Only five types of questions are used in this dataset, i.e.

*what is the <part>?*

*what are the <part>s?*

*where is the <part>?*

*where are the <part>s?*

*what can you do with your <part>(s)*

where *<part>* is the name of a body part.

### 2.4.3   The categorization dataset

The third dataset is used for evaluating the categorization capabilities of the system. This dataset uses 62 different animal names from 6 categories: 13 mammals, 13 birds, 13 fishes, 8 reptiles, 4 amphibians, 11 insects. The animal name memberships to the six categories are specified by 62 declarative sentences of the form:

*the <animal> is a <category>*

where *<animal>* is an animal name, and *<category>* is one of the six categories listed previously, as for instance:

*the turtle is a reptile*

Other 6 sentences of the form:

*<category>s are animals*

and one *how-to* sentence are included to train the system to deal with categorization hierarchies. The dataset also includes 48 declarative sentences of the form:

*the <animal> is <adjective>*

where *<adjective>* is one of the five adjectives: big, dangerous, domestic, fast or small. The total number of sentences in this dataset is 117.

In the training stage, the human teacher asks the system to tell him an animal belonging to one of the categories, e.g.

*tell me a mammal*

then he guides the system to a correct answer, as shown in detail in Appendix A, Sect. A.2. A single training example, involving one animal name from one category, is sufficient. After that, the system is able to answer correctly to the analogous question for all 6 categories. This test shows that the system is able to learn that the "is a" couple is used in sentences as "the dog is a mammal" to state that a concept belongs to a category, and that the "tell me a" group in a question can be used for asking to retrieve a concept from a category. A more complex categorization task in the same dataset involves the ability to learn categorization hierarchies. In this case, the human asks the system two consecutive questions, as in the following example:

Q: *what is the turtle?*
A: *it is an animal*
Q: *what kind of animal?*
A: *a reptile*

Other questions in this dataset are used to evaluate the capability of the system to combine information on categories and adjectives, as in the following example:

Q: *tell me a big reptile*
A: *crocodile*

### 2.4.4 The communicative interactions dataset

The fourth session is devoted to communicative interactions, and it is based on a mother/child dialogue extracted from the Warren-Leubecker corpus [43,44], which is part of the CHILDES database [45]. This corpus contains data from 20 children interacting with one of their parents.

The sessions took place in the child's home. The parent was instructed to bring the child into conversation and to talk to him as naturally as possible. This corpus appeared to be more appropriate than others for training the system, because the children ages were appropriate and because verbal communication was predominant over nonverbal communication, play and actions. The session used in this work is based on the file "david.cha", which contains a transcription of the dialogue between a 5-years-and-10-months-old child and his mother.

The system was trained in a text-based virtual environment. First, we guessed what kind of past experiences of the child could be compatible with the David dialogue: one day a relative brought the child to an amusement park; the child played to a video game (Pacman). Another day, at the kindergarten, the teacher organized a costume party, where each child should dress as a character that represents a letter of the alphabet. At home, the mother helped the child to prepare his letterman dress.

Those past experiences are described through a first set of 52 declarative sentences. Then we describe similar possible past experiences of the child impersonated by the system (a little girl, in our case): one day her father brought her, her sister and her cousin to the central park, were they played hide-and-seek and other games; another day, she was in Susan's room; aunt Carol told Susan to tidy up her room, therefore Susan started to put things inside her toy-box... Those experiences are described through another set of 44 declarative sentences, similar in syntax but different in the content from those of the first set. The training is based on this second set. Other 18 sentences in this dataset are *how-to* sentences. The human teacher guided the system into a conversation similar in syntax to the David dialogue, but related to a different past experience, and suggested either possible answers to the questions, or sentences appropriate for the conversation. In the test stage, the human interlocutor had a conversation with the ANNABELL system similar to that taken from the Warren-Leubecker corpus. Appendix B, Sect. B.2 shows a list of the informative sentences used to build the system experience in a virtual text-based environment. Appendix B, Sect. B.3 shows the sentences used to train the system.

### 2.4.5 The virtual environment dataset

The fifth dataset represents a text-based virtual environment, where the system is trained to perform simple tasks by means of verbal commands. The training is made in a virtual house with 25 room, named *room_0*, …, *room_24*, arranged in a 5×5 square. A person is located in the central room, i.e. *room_12*, which is also the starting position of the system. Eight objects, named *object_1*, …, *object_8*, are distributed randomly in the eight second-nearest-neighbor rooms, with the constraint that different objects should be located in different rooms. Each room has a description that specifies the accessible nearest neighbors and eventually the persons and/or objects that are present in it, as

*you are in room_12*

*to the east there is room_13*

*to the north there is room_7*

*to the west there is room_11*

*to the south there is room_17*

*John is here*

*what do you want to do?*

Before the training, the descriptions of all rooms are presented to the system. Each training example starts by presenting to the system the description of the starting room (*room_12*) and by asking it to accomplish the task of bringing an object to the person:

*bring an object_5 to John*

then, the system is trained to issue the commands (*north*, *south*, *east*, *west*) to move to the room where the object is located. Each time the system moves to a room, it receives the description of that room. When it reaches the target room, it is trained to take the object

*take the object_5*

to go back to the starting room, and to give the object to the person

*give the object_5 to John*

The test is made in the virtual house shown in Fig 6.

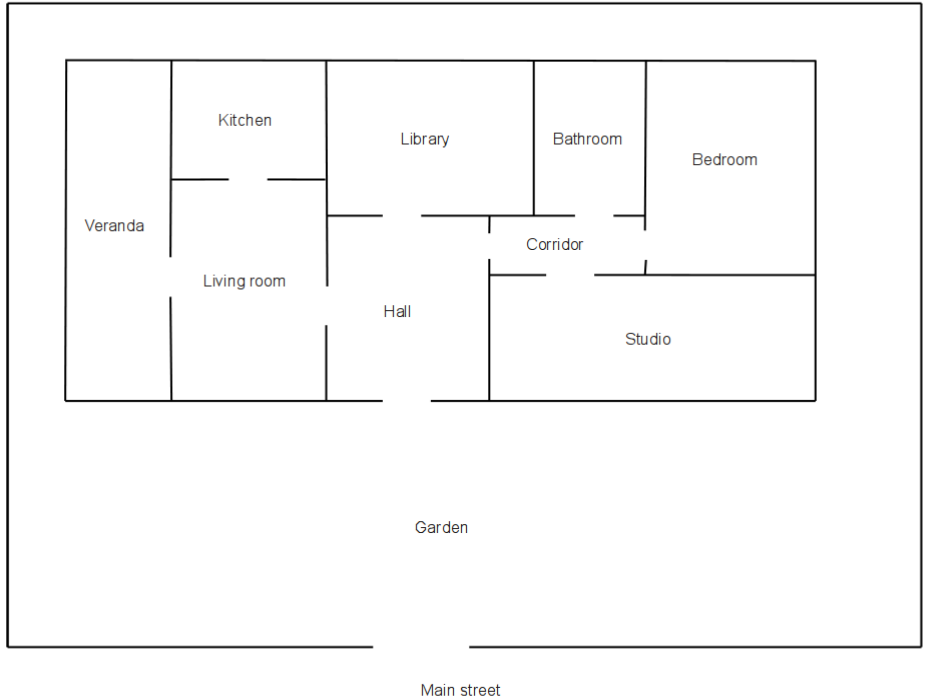

**Fig 6.** Map of the virtual house used to build the sentences for the test set of the *virtual-environment* dataset.

Before starting it, the descriptions of all rooms are presented to the system. All possible combinations of starting room and target room are used in the test, using for simplicity the constraint that starting room and target room are second nearest neighbors. The number of combinations for this house is 28. For each combination, the system and a person are located in the starting room, and an object (a book) is located in the target room. After the description of the starting room, the system is asked to bring the object to the person:

*bring a book to Alfred*

As in the training stage, in order to perform this task, the system should issue the commands for reaching the target room, the command for taking the object, the commands for going back to the starting room, and the command for giving the object to the person.

# 3  Results

The training procedure is organized in five incremental language training sessions, one for each dataset. Each session is divided in two stages. During the first stage, a set of declarative sentences from the corresponding dataset is presented to the system through the interface. As the system does not have any other sensory input, all the information must be provided to it in the form of verbal descriptions. In the subsequent training stage, the teacher trains the system by asking it a set of questions related to the previous sentences, and by guiding it to produce the correct answers through the exploration-reward procedure described in Sect. 2.3.

The evaluation of the system performance (test stage) is performed at the end of the five learning sessions, after the cumulative training on all five datasets. In this stage, the teacher evaluates the system by asking it a set of questions similar to the ones used during the training stages, and by testing the generalization capabilities of the system, i.e. its ability to process the information provided by the memorized sentences, and to answer questions having a similar structure to those presented during the training stages but involving different nouns, adjectives or verbs. The teacher also validates the linguistic competences of the system in the use of articles, nouns, verbs, adjectives, personal pronouns, possessive pronouns and other word classes. The system output sentences were considered to be valid when they were syntactically and semantically correct and appropriate for the conversation. The test related to the virtual environment dataset evaluates whether the system is able to generalize the knowledge acquired in the training stage, being able to follow similar commands involving different target rooms, objects, people.

Table 3 reports the number of declarative sentences, the number of interrogative sentences used for training, the number of interrogative sentences used for the test and the number of output sentences in the five datasets.

**Table 3.** Number of declarative sentences, number of interrogative sentences used for training, number of interrogative sentences used for the test and number of output sentences in the five learning sessions. The conversational sentences that do not expect a turn taking from the system are treated as declarative sentences, while the ones that expect a turn taking are treated as interrogative sentences. The output sentences in the virtual environment (*) are actually commands issued by the system to perform actions in the environment itself.

| Session | Declarative sentences n. | Interrogative sentences n. (training) | Interrogative sentences n. (test) | Output sentences n. (test) |
|---|---|---|---|---|
| People | 225 | 68 | 284 | 270 |
| Parts of the body | 122 | 16 | 75 | 74 |
| Categorization | 117 | 8 | 144 | 142 |
| Communicative interactions | 114 | 32 | 29 | 35 |
| Virtual environment | 167 | 18 | 168 | 168(*) |
| **Total** | **745** | **142** | **700** | **521+168(*)** |

Fig 7 shows how the the interrogative sentences of the first four datasets used in the training stages and those used in the test stages are distributed among the question categories. The virtual environment dataset was excluded from this statistic because in this case the interrogative sentence is always the same, i.e. it is the question "*what do you want to do?*", which is used to ask the system what action it wants to perform.

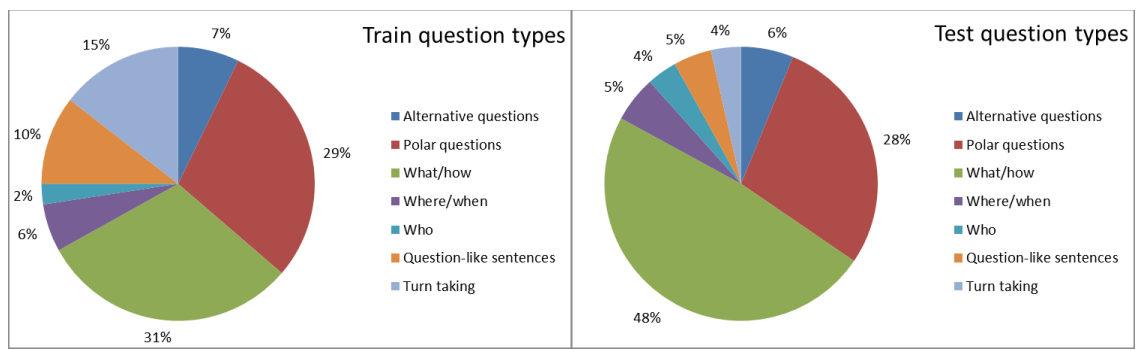

(a) (b)

**Fig 7.** Distribution of the questions used for training (a) and for test (b) among the question categories: alternative (choice) questions, polar (*yes/no*) questions, *what/how* questions, *where/when* questions, *who* questions, question-like imperative sentences (e.g. *tell me*), conversational sentences that expect a turn taking from the system.

In order to evaluate quantitatively the system performance, we used a four-rounds cross validation approach. The communicative interaction dataset was excluded, as it was not suitable for a cross validation. All the questions of the first three datasets were organized in groups, each group containing at least four interrogative sentences having similar structure. At each round, the training set was built by randomly extracting one or more questions from each group, with the constraint that the same question should not be used in different rounds. The remaining questions were used for the test. The order of the questions used for training and that of the questions used for the test were both randomized.

Concerning the virtual environment dataset, the four rounds of the cross validation used different starting seeds for extracting randomly the position of the target rooms used in the training stage. The test was always made in the virtual house shown in Fig 6, on all possible combinations of starting room and target room.

For each round of the cross validation, the system was first trained on all five datasets before testing it. In this way we could test the capabilities of the system to store all the information of the five datasets and to acquire new information without altering the past one.

In a single round of the cross validation, the system was trained and evaluated using 1587 input sentences, containing 595 different words, with an average number of 5.6 words per sentence. It produced 521 output sentences, containing 312 different words (expressive vocabulary), with an average number of 4.6 words per sentence. Fig 8 shows the distribution of the number of tokens (words) in the input sentences (a) and in the output sentences (b), the distribution of different tokens among word classes (c,d), and the percentage of word classes in the input and output sentences (e,f).

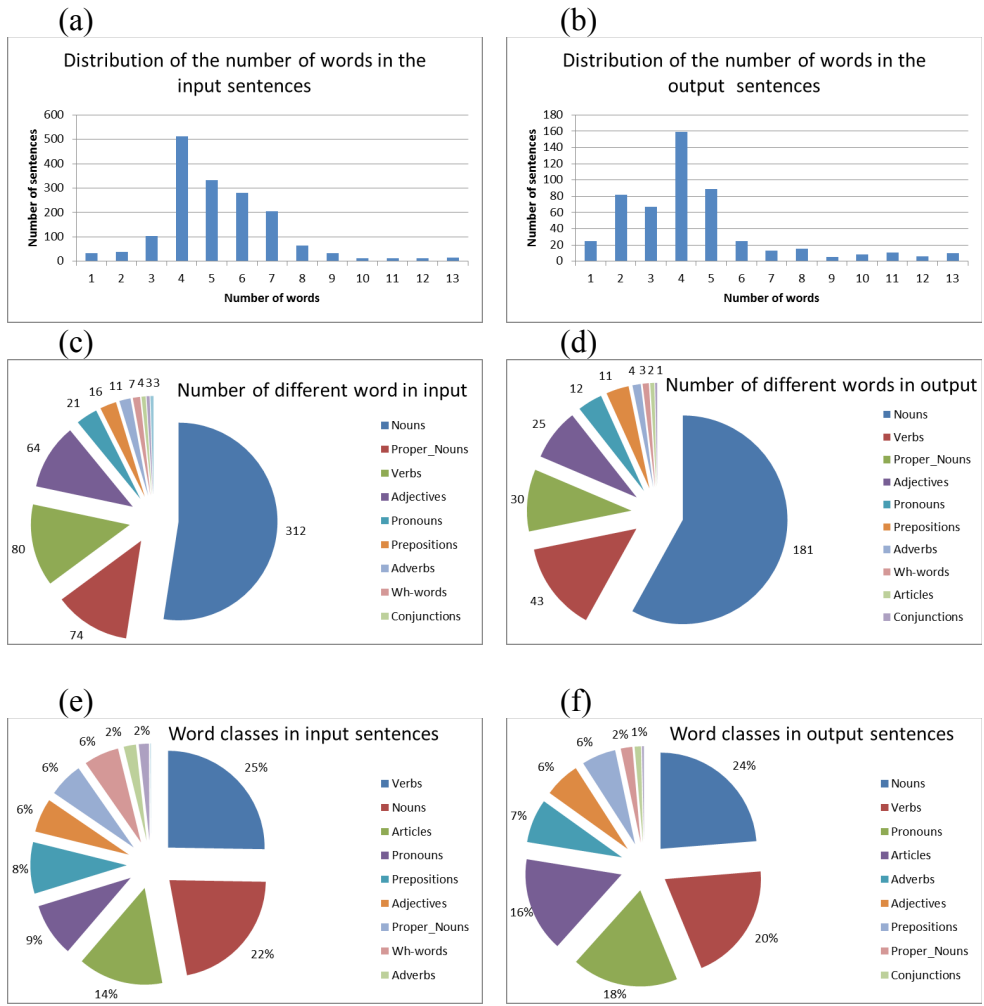

**Fig 8.** Distribution of the number of words and of the word classes in the input and output sentences. Distribution of the number of words in the input sentences (a) and in the output sentences (b); distribution of the words used in the input and output sentences among different word classes (c,d); percentage of word classes in the input and output sentences (e,f).

Table 4 reports the number of correct answers for the first three datasets and for the four rounds of the cross validation. Table 5 reports the number of tasks that were performed correctly by the system on the virtual environment dataset for the four rounds of the cross validation as a function of the number of training examples.

**Table 4.** Number of correct answers over the total number of expected answers in the test stage of the cross-validation rounds for the first three datasets. An answer to an interrogative sentence is considered to be valid only if it is correct both syntactically and semantically.

| Dataset | Round 1 correct/total | Round 2 correct/total | Round 3 correct/total | Round 4 correct/total | Average correct/total (%) |
|---|---|---|---|---|---|
| People | 235/284 | 224/284 | 239/284 | 238/284 | 82.4 % |
| Parts of the body | 64/75 | 65/75 | 65/75 | 62/75 | 85.3 % |
| Categorization | 140/144 | 139/144 | 136/145 | 136/145 | 95.3% |

**Table 5.** Number of tasks that are performed correctly by the system on the virtual environment dataset over the total number of assigned tasks in the test stage of the cross validation rounds, as a function of the number of training examples used in the training stage.

| Number of training examples | 1 | 2 | 3 | 4 | 5 |
|---|---|---|---|---|---|
| Round 1 correct/total | 18/28 | 22/28 | 27/28 | 28/28 | 28/28 |
| Round 2 correct/total | 24/28 | 24/28 | 24/28 | 28/28 | 28/28 |
| Round 3 correct/total | 24/28 | 24/28 | 24/28 | 24/28 | 28/28 |
| Round 4 correct/total | 27/28 | 28/28 | 28/28 | 28/28 | 28/28 |
| Average correct/total (%) | 83.0 | 87.5 | 92.0 | 96.4 | 100 |

The percentage of correct output sentences over the total number of requested output sentences, averaged over the four rounds of the cross validation, was 82.4% for the people dataset, 85.3% for the parts of the body dataset, and 95.3% for the categorization dataset. The communicative interaction dataset is excluded here because it is not suitable for this type of quantitative evaluation. The average percentage of tasks that were performed correctly by the system in the virtual environment over the total number of assigned tasks varies from 83.0% to 100% depending on the number of training examples used in the training stage, as shown in Table 5.

In the test related to the communicative interaction dataset, the human interlocutor brought out the system in a conversation similar to the one transcribed in the corpus. Fig 9 shows an extract of a side-by-side comparison between the human/ANNABELL-system conversation on one side and child/parent interaction on the other side. The complete comparison is reported in Appendix B, which also reports the sentences used for the training.

| SYS: ANNABELL system; TEA: human | CHI: real child; MOT: Mother |
|---|---|
| *TEA: Ann, why do not you tell, tell Dad about you playing Pac_Man . | *MOT: Christopher , why don't you tell , tell Daddy about you playing Pac_Man . |
| *SYS: Michael brought me to the bumpers junction . | *MOT: didn't Michael take you to play Pac_Man at bumper's junction ? |
| *SYS: I played Pac_Man . | *MOT: huh ? |
| *TEA: did you win ? | *CHI: yeh . |
| *SYS: yes, I won . | *MOT: well ? |
| *TEA: did you play other games ? | *MOT: did you win ? |
| *SYS: I played Space_Invaders . | *MOT: what did you do ? |
| *TEA: did you win ? | *CHI: I won ! |
| *SYS: no, I did not win . | *MOT: how many games did you play ? |
| *TEA: how many games did you play ? | *CHI: &um , I don't know . |
| *SYS: I played : | *MOT: you don't know ? |
| *SYS: Space_Invaders, one ; | *MOT: I know what's better to talk about . |
| *SYS: Pac_Man, two ; | *MOT: you need to tell (th)em about , &um , junk+man . |
| *SYS: Donkey_Kong, three ; | *CHI: xxx . |
| *SYS: three games . | *MOT: your letter man ! |
| *TEA: I know what is better to talk about . | *CHI: no ! |
| *TEA: you need to tell them about jumble junkman . | *MOT: yes ! |
| *SYS: jumble junkman will be my letter man . | *MOT: tell (th)em what I put all over your shirt . |
| *TEA: how about Harry hair ? | *MOT: &um &um . |
| *SYS: Harry hair will be Bob's letter man . | *CHI: sticker . |
| *TEA: tell me what Mum put on your shirt . | *MOT: sticker . |
| *SYS: a sticker . | |

**Fig 9**. Extract of a side-by-side comparison between the human/ANNABELL-system dialogue on one side and the mother/real-child dialogue on the other side, based on the Warren-Leubecker corpus from the CHILDES database. The right side is a transcription of a conversation between a 5-years-and-10-months old child and his mother, extracted from the file "david.cha" of the CHILDES database. Note that the human/ANNABELL dialogue system does not use punctuation, which has been added here for clarity.

Fig 10 shows the distributions of the number of tokens in the output sentences of our system and in the child utterances, from the session based on the Warren-Leubecker corpus. The total number of tokens used by the child was 134, while those used by the system in this session were 111. The total numbers of different token types were 86 and 75, respectively. The type/token ratios are close to each other, being 0.64 for the child, and 0.68 for our system (this analysis was performed using the CLAN program [45] ).

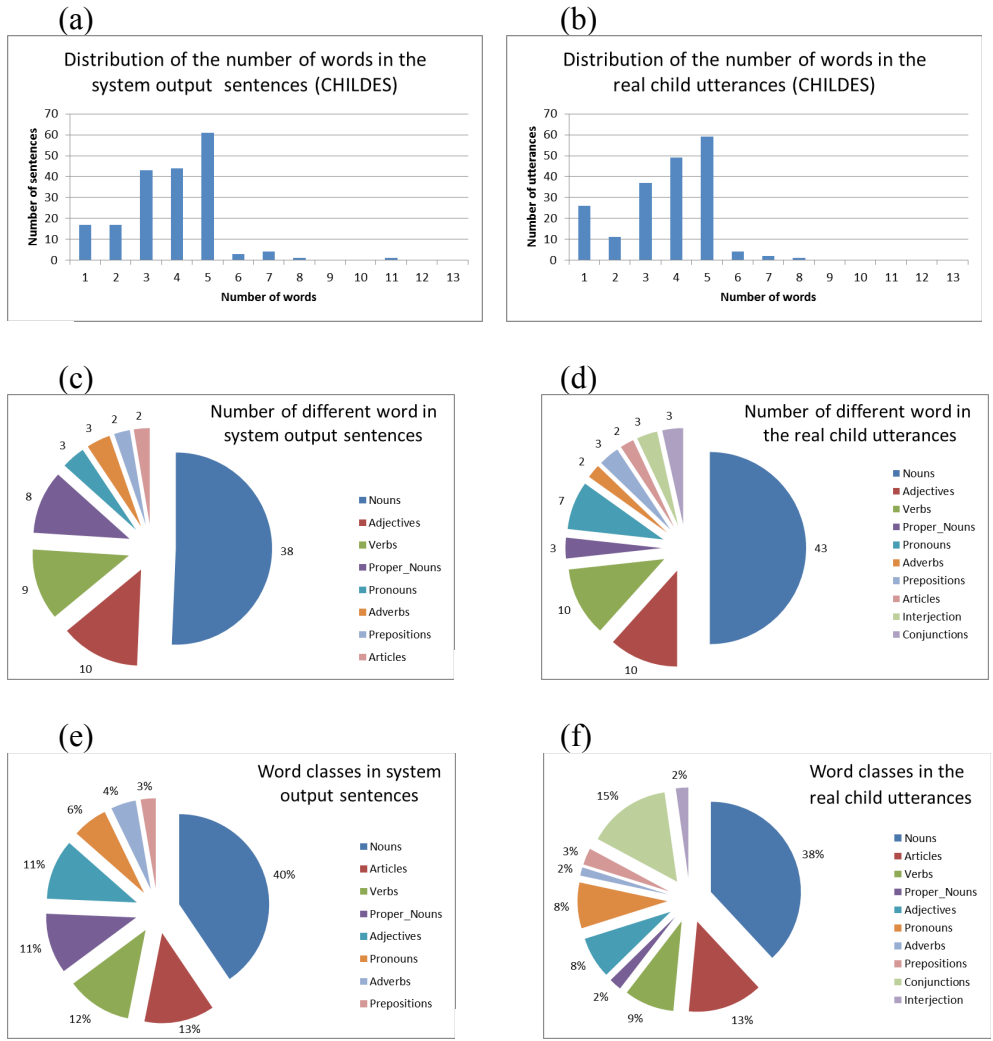

**Fig 10.** Comparison between some distributions related to the output sentences produced by the system in the communicative interaction test, based on the Warren-Leubecker corpus from the CHILDES database, and to the utterances of the real child for the same part of the corpus: distribution of the number of words in the output sentences (a,b); distribution of the words used in the output sentences among different word classes (c,d); percentage of word classes in the output sentences (e,f).

Fig 11 shows how the average time that the system needs to answer to a question varies with the number of allocated input connections in the state-action association subnetwork. At the end of the whole training process, performed on all five datasets, the average time for an answer, evaluated on a system equipped with a high-performance GPU (graphics processing unit) NVIDIA Kepler GK104 having 1536 cores, is 9.5 seconds.

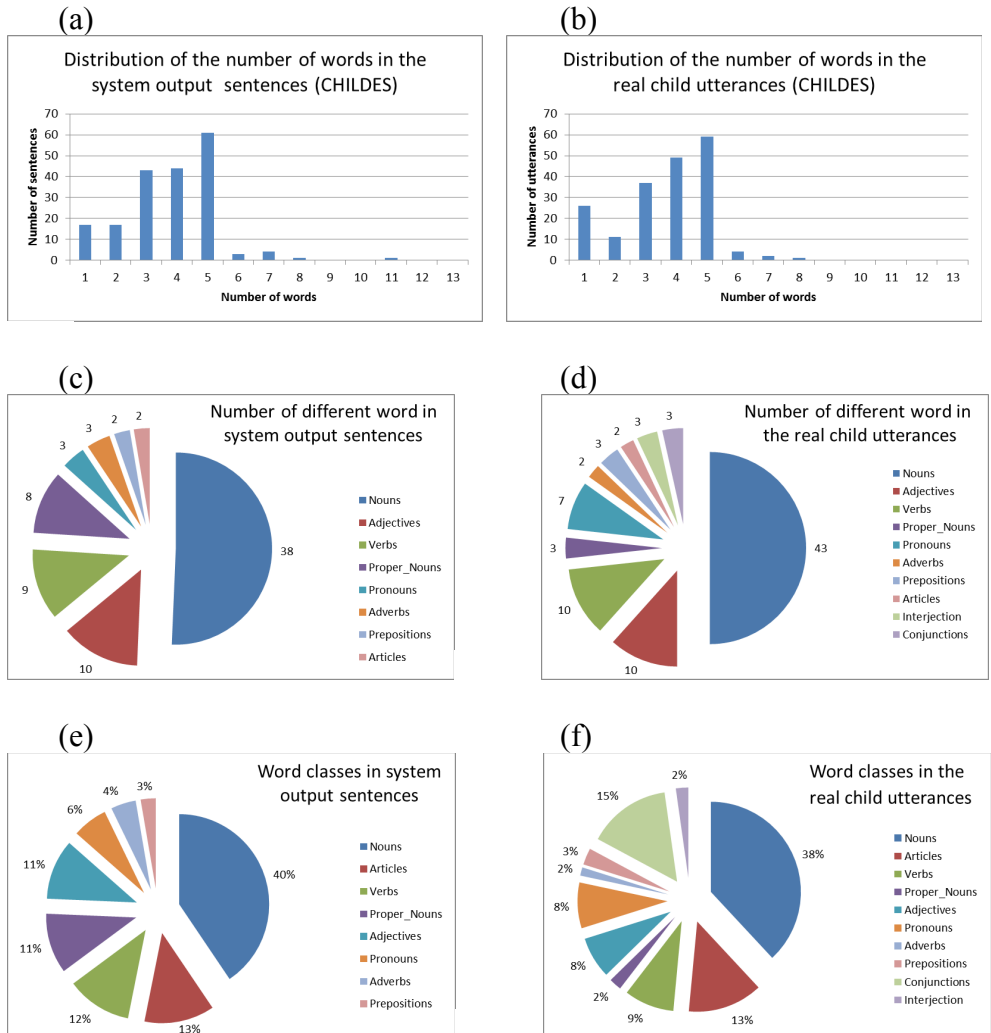

**Fig 11.** Answer time. Average time that the system needs to answer to a question as a function of the number of allocated connections in the state-action-association subnetwork, evaluated on a CPU system based on a Xeon 2.50 GHz dual-processor-quad-core 16GB RAM, and on a system equipped with a high-performance GPU NVIDIA Kepler GK104 having 1536 cores.

## 3.1 Evaluation of the system components and free-parameters optimization

The STM of the ANNABELL model is organized into different components, as described in Sect. 2.3. All these components are connected to the central executive, which uses their neural activation states for the decision processes that associate mental actions to the internal states of the system. In this section we discuss the relative importance of each component for this decision process, and the effect of their complete removal from the system.

Table 6 shows how the system performance is affected by a removal of the connections from the STM components to the central executive, and how it is affected by a complete removal of each of these components from the system. The percentage of correct answers is averaged over the three datasets people, parts of the body and categorization, and over the four rounds of the cross validation.

**Table 6.** Percentage of correct answers after removal of the connections from the STM components to the central executive, and after complete removal of the components. The values are averaged over the three datasets people, parts of the body and categorization, and over the four rounds of the cross validation. The percentage of correct answers with all components and with all connections is 86.5 %.

| Component | Removal of connections to the CE (correct/total) | Complete removal of the component (correct/total) |
|---|---|---|
| Comparison structure | 76.0 % | 76.0 % |
| Goal stack | 81.2 % | 66.5 % |
| Previous phrases | 82.3 % | 82.3 % |
| Indexes | 83.8 % | 0 % |
| Input-phrase buffer | 79.8 % | 0 % |
| Current word | 85.1 % | 0 % |
| Working-phrase buffer | 43.6 % | 0 % |
| Word-group buffer | 86.4 % | 0 % |
| Output phrase buffer | 86.5 % | 0 % |

The contents of the STM components are not independent of each other: there is a redundancy in the information that they provide. Therefore, the removal of the connections from a single component to the central executive does not compromise the system functionality completely. The previous-phrases structure and the comparison structure are only used to provide an input to the state-action association system, so their complete removal has the same effect as the removal of their connections to the central executive. It may be noted that this removal produces a decrease in the percentage of correct answers. If the goal structure is completely removed from the system, the processes of insertion and extraction of the phrases in the goal stack will have no effect. The system can still work but since it has lost an important feature, its performance will decrease substantially, as can be observed in Table 6.

The other components of the STM are essential to the functionality of the system, as described in Sect. 2.3. The system can not work properly if they are completely removed.

Fig 12(a) and 12(b) show the percentage of correct answers as a function of the parameter $W_{max}$ of the DHL rule, used to update the connections from the STM components to the central executive.

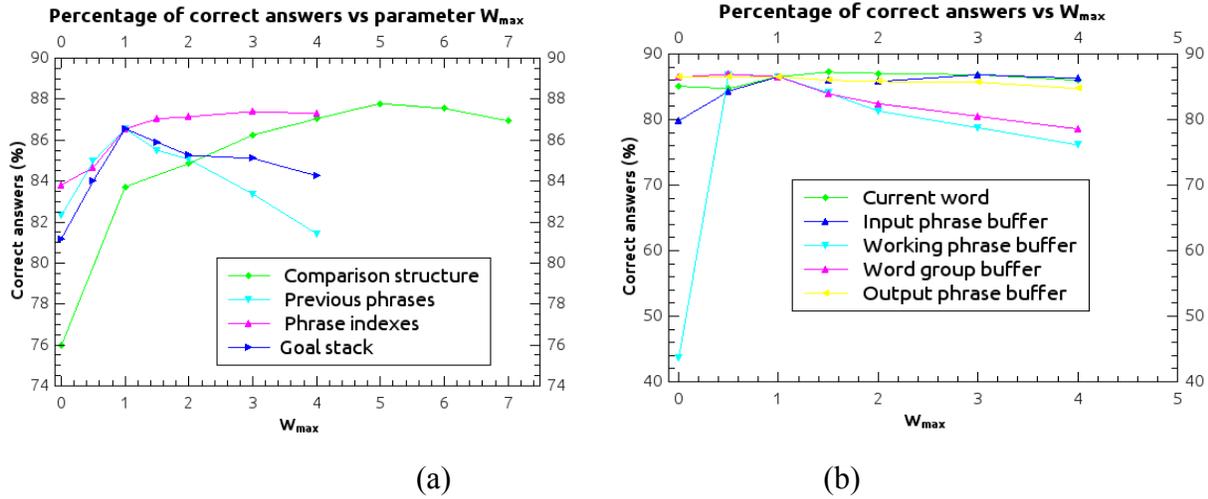

(a)　　　　　　　　　　　(b)

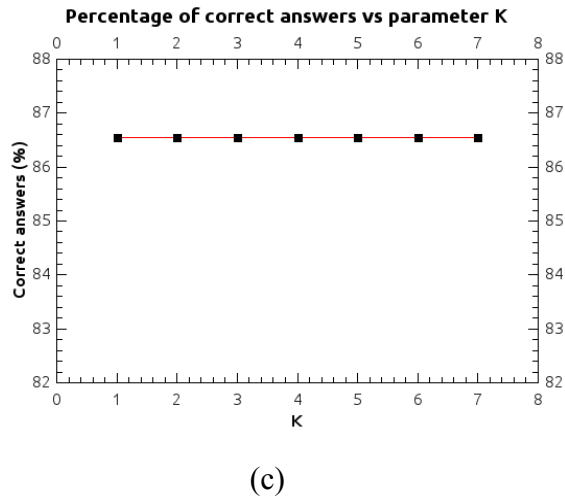

(c)

**Fig 12.** Percentage of correct answers over the total number of expected answers, evaluated on the first three learning sessions: [(a) and (b)] as functions of the weight-saturation value $W_{max}$ for the main components of the system; (c) as a function of the parameter $k$ used for the $k$-winner-take-all algorithm in the state-action-association subnetwork.

$W_{max}$ is the weight-saturation value. A null value of $W_{max}$ has the same effect as a removal of the connections from a STM component to the central executive. A variation of $W_{max}$ produces a change in the relative weight of the component in the decision process operated by the central executive, as will been shown in Appendix D. With the exception of the comparison structure, it can be observed from Figs 12(a) and 12(b) that all plots have a maximum in the range $0.5 \leqslant W_{max} \leqslant 3$, and that the variations of the system performance are relatively small in this range. On the other hand, the plot for the comparison structure has a maximum for $W_{max} \simeq 5$. In this work we have used the same value ($W_{max}$=1) for all STM components, except for the comparison structure, for which we used the optimal value.

Fig 12(c) represents the percentage of correct answers as a function of the parameter *k* used for the k-winner-take-all rule in the state-action-association system. It can be observed that this percentage does not change with the value of *k*. This is related to the fact that the system uses a discrete version of the Hebbian learning rule. The value of $W_{max}$ for the comparison structure and the value of *k* are the only two free parameters used in our model.

## 3.2  Generalization

Our study is focused on the children age range between about 3 to 5 years, which is a crucial range for the acquisition of linguistic competencies, and therefore is considered particularly interesting for studies on language development. The sentences in the databases described in Sect. 2.4 have been chosen according to the purpose of this work, based on the literature on early language assessment [41-45]. For such reason, the grammatical structure of the sentences in the datasets described in Sect. 2.4 is relatively simple. Nevertheless, it is important to evaluate the generalization abilities of the system on a larger dataset and on more complex grammatical constructions. We can distinguish two types of generalization [21]:

1) handling learned grammatical constructions with new open-class words;

2) compositional generalization, i.e. generalize knowledge to new constructions that were not used in the training corpus.

In the following paragraphs, we evaluate the system performance in two experiments related to these two types of generalization.

### 3.2.1 Generalization 1

For this experiment, we built an extended database by replacing the open-class words of the three datasets *people*, *parts of the body* and *categorization*, with new, randomly generated words.

The declarative sentences of this database were produced by using the constructions shown in table 1 for the *people* dataset, the constructions of table 2 for the *parts-of-the-body* dataset and the sentences described in Sect. 2.4.3 for the *categorization* dataset; the open-class words within the angle brackets of those constructions have been replaced by randomly generated words, i.e. random alphabetical strings. The open-class words in the interrogative sentences used for the test were modified accordingly. By iterating this procedure, we generated a database of 5352 declarative sentences and 4028 interrogative sentences. This database was used as an independent test set for testing the four instances of the system that were trained on the original database during the four rounds of the cross-validation, respectively.

Table 7 shows the number of correct answers produced by the four instances of the system over the number of interrogative sentences for the three extended datasets people, parts of the body and categorization. The results demonstrate that the system is able to generalize the acquired knowledge to learned construction with different open-class words.

**Table 7.** Number of correct answers produced by the four instances of the system over the number of interrogative sentences for the three extended datasets people, parts of the body and categorization. The four instances of the system are the ones obtained by training the system on the original datasets during the four rounds of the cross-validation, respectively.

| Dataset | Instan885/1156ce 1 correct/total | Instance 2 correct/total | Instance 3 correct/total | Instance 4 correct/total | Average correct/total (%) |
|---|---|---|---|---|---|
| People | 2008/2272 | 1994/2272 | 1984/2272 | 2021/2272 | 88.1 |
| Parts of the body | 537/600 | 520/600 | 535/600 | 527/600 | 88.3 |
| Categorization | 885/1156 | 887/1156 | 887/1156 | 887/1156 | 76.7 |

### 3.2.2 Generalization 2

The compositional generalization capacity of the system was evaluated through an experiment of sentence-to-meaning mapping, based on a task that was developed by Caplan et al. [46].

In the Caplan task, an aphasic subject listens to sentences and then he is required to indicate the meaning by pointing to images depicting the agent, object and recipient, always in that canonical order.

In formal terms, the input is the sequence of words in the sentence, and the output is the sequence agent, object, and recipient, corresponding to a standardized representation of the meaning in terms of thematic role assignment.

In our implementation, the surface form of the input sentences is presented word-by-word to the system, which is trained to assign the the thematic roles (predicate, agent, object, recipient) of the open-class words. For this experiment we used a dataset of 462 distinct grammatical constructions developed by Hinaut and Dominey [22], who used a context free grammar to generate systematically distinct grammatical constructions, each consisting of between 1 and 6 nouns, with 1 to 2 levels of hierarchical structure (i.e. with only a main clause or a main and relative clause, respectively). Each grammatical construction of this dataset has a surface form and a coded meaning. The surface form is composed by the word-groups shown in table 8.

**Table 8.** Group of words that compose the sentences of the database used for this experiment. The X represents a closed-class word. In case of ambiguities, the system is trained to use the largest groups.

| X | X -s | X -ed | that |
|---|---|---|---|
| the X | was | X -ed it | - |
| to the X | was X -ed | X -ed was | - |
| by the X | was X -ed by | X -ing was | - |

Our model operates in two stages:

1) grouping: the system is trained (on the training set) to split the input sentence in word groups, according to table 8, and to send each word group to the output. In case of ambiguities, the system is trained to use the largest groups. After the whole sentence is split, the output phrases are fed back to the system as new input phrases.

2) thematic role assignment.

The architecture of our model does not include a structure where the open-class words can be explicitly mapped to their thematic role. In order to perform this task without modifying the system architecture, our approach was to explicitly ask the system for the thematic roles:

? predicate

? agent

? object

? recipient

? relative-clause predicate

? relative-clause agent

? relative-clause object

This approach is more close to the Caplan-task protocol. It should also be noted that our model, in contrast to other approaches to the same problem, does not require a prior specification of the distinction between open-class and closed-class words.

Following the same approach of Ref. [22], the compositional generalization capacity of our model was tested in a ten-fold cross-validation. The dataset was divided in ten partitions (eight partitions with 46 sentences, and two with 47 sentences). In each round of the cross-validation, the system was trained on nine partitions and tested with the one not used for training. This procedure was performed ten times, so that all partitions were used for the test. Table 9 reports the results of the cross validation. *Meaning error* is the percentage of incorrect thematic role assignment. *Sentence error* is the percentage of sentences in which there is at least one wrong thematic role assignment. As illustrated in the table, the cross validation yielded 9.2% average meaning error and 36.7% average sentence error rates. The model proposed by Hinaut and Dominey achieved 9.2% average meaning error and 24.4% average sentence error rates through a ten-fold cross validation on the same corpus. This means that the number of errors in thematic role assignment is the same, however in their work the assignment errors are concentrated in a smaller number of sentences. It should be considered that while that work is focused on the problem of thematic role assignment, our model is not optimized for this specific task, because it addresses a wider range of aspects of human language processing.

**Table 9.** Meaning errors and sentence errors on the ten rounds of the ten-fold cross validation. *Meaning error* is the percentage of incorrect thematic role assignment. *Sentence error* is the percentage of sentences in which there is at least one wrong thematic role assignment.

|  | Round 1 | Round 2 | Round 3 | Round 4 | Round 5 | Round 6 | Round 7 | Round 8 | Round 9 | Round 10 | Average |
|---|---|---|---|---|---|---|---|---|---|---|---|
| meaning error rate | 11.1% | 10.8% | 8.5% | 9.3% | 11.0% | 10.3% | 9.5% | 5.8% | 8.4% | 6.8% | 9.2% |
| Sentence error rate | 42.6% | 38.3% | 34.8% | 39.1% | 39.1% | 47.8% | 43.5% | 23.9% | 32.6% | 26.1% | 36.7% |

## 4  Discussion

Several examples in the test show that the ANNABELL system expresses a broad range of capabilities compared with other cognitive neural systems. For instance, let us consider the age-comparison example, described in Sect. 2.4.1, or the question "How many games did you play?" in Appendix B from the CHILDES database. The answers to such questions involve a process quite more complex than simple information retrieval from LTM and rearrangement of input and retrieved sentences. The first example involves counting skills, ability to compare small numbers, ability to associate the words "your friend" to a known person, ability to retrieve information about her age from the LTM, ability to use personal pronouns. The system is able to learn how to answer to this question through a rewarding procedure, and to generalize the acquired knowledge to similar questions involving different people with different ages. Importantly, our system does not include a specialized structure for counting, or a specialized structure for number comparison, or a specialized structure for mapping names into personal pronouns.... All its abilities arise from a relatively small set of mental actions, that allow it to extract a word-group from the phrase stored in the phonological store, to use this word-group as a cue for retrieving other phrases from LTM, to insert and retrieve phrases from a goal stack.

In the second example ("How many games did you play?") the system is able to retrieve the three games from LTM and to count them. It is important to point out that the system performs such tasks through sequences of mental operations that are compatible with psychological findings. Even though our neural model is extremely simplified (compared to biologically realistic neural models), it can help to understand the link between neural activity and large-scale organization described by theoretical models. Previous neural models of language lack a central executive, or some other system that controls the flow of information among the other STM subsystems. We propose that this control can be done by neural gating mechanisms. This hypothesis is compatible with recent research, which demonstrates that neural gating mechanisms play a fundamental role in the flow of information in the cortex. We also provide a model of how the central executive can learn how to associate mental actions to the internal states of the system through a rewarding procedure.

In the past decades, many researchers emphasized the contrast between localist and distributed models. In our work, word representation is based a localist model. On the other hand, the central executive, which is the heart of our system, follows a distributed model, and our work emphasizes that the decision processes operated by this component are not based on pre-coded rules, but statistical. It is important to point out that word representation is not a central point of our work. The localist representation of words can be regarded as a simplification, mainly motivated by the need for computational efficiency. Conversely, the use of a sparse signal representation is a basic feature of our model. In principle, our model could be modified to use a distributed (but still sparse) representation of the words. The system would have a better ability to recognize similarities (semantical and/or phonological) among words, at the expense of a much greater number of connections and thus a much larger computation time. This could be a subject for a future work.

Beyond the debate on word representation, there are several points of novelty of our model compared to previous cognitive models of language:

- it is the first system entirely based on an artificial neural network that is able to sustain a dialog, expressing a broad range of functionalities in language processing.
- It provides a working neural model for the process of "inner though" that occurs between language acquisition and language production. This model is compatible with the multicomponent working memory framework, which is supported by a large number of findings from neuroimaging and psycholinguistic.
- It processes the information through mental actions, which are controlled by a central executive and are performed by neural gating mechanisms.

In the context of human-computer interaction, human language understanding is often associated to the ability to translate a linguistic input into a standardized functional form. This type of understanding involves the capacity to recognize the thematic role of the open-class words in the surface form of sentences. Meaning in this case is interpreted as a mapping from the surface form to the functional form. Our model does not have a structure where this mapping is explicit, however its ability to identify thematic roles can be tested through a question-answer approach, as in the Caplan task that was discussed in section 3.2.2.

The previous notion of understanding is insufficient for the purpose of our work. Question answering and, more generally, communicative interactions involve a kind of procedural knowledge, which is used to process the linguistic input and to produce the output. This type of understanding refers to the ability to perform the sequences of mental operations that are needed to respond to a verbal input. For instance, let us consider the input sentence "Tell me about your classmates". A response to this sentence requires understanding that the words "your classmates" refer to a set of individuals, extracting from LTM information about those individuals, selecting part of this information and processing it in a form useful for the output sentences. Our work is an attempt to implement a working cognitive model that helps to understand the development of this procedural knowledge.

Many researchers argued that a true understanding can not be achieved if language is not grounded in the agent's physical environment through actions and perceptions [47]. An active field of research is devoted to grounding open-class words to objects (visual elements, bodily sensations and other types of perceptions) and grounding sentences to scenes and actions. Morse et al. [48], Cangelosi and Schlesinger [49] highlighted the role of embodiment in early language development. Dominey and Boucher [50] argued that we learn to translate the surface form of language into a functional form through the integration of speech inputs and non-speech inputs. In Baddeley's working memory model, this integration occurs in the episodic buffer. A limit of the current version of our model is that language is not grounded. Language grounding would require the combination of our model with a visual system, or its embodiment in a larger system that integrates language with other forms of perceptions and actions.

## 5   Conclusion

The results of the validation show that, compared to previous cognitive neural models of language, the ANNABELL model is able to develop a broad range of functionalities, starting from a *tabula rasa* condition. The system processes verbal information through sequences of mental operations that are compatible with psychological findings. Those results support the hypothesis that the central executive plays a fundamental role for the elaboration of verbal information. Our work emphasizes that the decision processes operated by the central executive are not based on pre-coded rules. On the contrary, they are statistical decision processes, which are learned by exploration-reward mechanisms. The reward is based on Hebbian changes of the learnable connections of the central executive. A neural architecture is suitable for modeling the development of the procedural knowledge that determines those decision processes.

The current version of the system sets the scene for subsequent experiments on the fluidity of the brain and its robustness in the response to noisy or altered input signals. Moreover, the addition of sensorimotor knowledge to the system (e.g. visual input and action capabilities) would lead to the extension of the model for handling the developmental stages in the grounding and acquisition of language [48].

**Acknowledgments:** This work was partially supported by the Regione Autonoma della Sardegna with funds from Operative Program FSE 2007–2012 L.R.7/2007 "Promozione della ricerca scientifica e dell'innovazione tecnologica in Sardegna". Cangelosi's efforts were funded as part of the UK EPSRC project BABEL and the FP7 Projects POETICON++. We are deeply grateful to Prof. Risto Miikkulainen for his invaluable discussions and suggestions for the improvement of our work.

# Appendix A  Examples from the learning sessions

The following conventions are used when typing phrases and/or commands through the interface:

- only lowercase letters are used, with no punctuations, no special characters; uppercase letters are used only for the first character of proper nouns; sentences do not start with a capital letter (unless they start with a proper noun);
- by convention, questions starts with a question mark, as: "*? how old are you*";
- *w*ords with suffix are split in the form: base -suffix; e.g. animals → animal -s, writing → write -ing, etc.

## A.1  First example: verbs and personal pronouns

In this example the system should combine the use of some verbs with that of personal pronouns. The teacher can start by typing the two phrases:

*the personal pronoun for a male person is he*

*the personal pronoun for a female person is she*

then he should type phrases as

*Susan is a female name*

*Susan is a doctor*

*Susan is drive -ing the car*

*Tim is a male name*

*Tim is a student*

*Tim is read -ing a book*

*Elisabeth is a female name*

*Elisabeth is a secretary*

*Elisabeth is write -ing a letter*

*Max is a male name*

*Max is an actor*

*Max is go -ing to the theater*

…

and other similar phrases with different names. The order in which the phrases are submitted is not important, and they can be mixed with other kinds of phrases.

Now the teacher can ask the question

*? what is Max do -ing*

starting with a question mark, without other punctuations, and following the rule discussed previously for words with suffixes. Then he can suggest target word groups and target phrases that lead to the correct answer. Since he would like that the system uses the personal pronouns, the first part of the output should be the word "he", which can be obtained through the following word-group extractions and associations:

*.word_group Max*

*.phrase Max is a male name*

*.word_group male*

*.phrase the personal pronoun for a male person is he*

*.word_group he*

The word group obtained at the end of this exploration phase is the first part of the output. It should be rewarded, however the system should be warned that the output phrase is not complete. The command that the teacher should use in this case is

*.partial_reward*

The only difference between the partial reward and the complete reward is that the first is terminated by a CONTINUE action, while the latter is terminated by a DONE action.

To complete the answer, the teacher can suggest the following word-group extractions and associations:

*.word_group Max*

*.phrase Max  is go -ing to the theater*

*.word_group is go -ing to the theater*

The last word group is the second and final part of the desired output, therefore it should be rewarded:

*.reward*

Now the teacher can test if the system is able to answer to similar questions, as for instance:

*what is Tim do -ing*

*.exploitation*

The system will answer:

*he is read -ing a book*

Note that it is not necessary to train the system with an example using a female name: the system will be able to use correctly both personal pronouns according to the gender. For instance, if the teacher asks the question:

*? what is Elisabeth do -ing*

the system will answer correctly, being able to use correctly both personal pronouns according to the gender.

## A.2  Second example: categorization

This example uses the animal classification to show the categorization ability of the system. A first very simple test can be made by launching the program and typing phrases as:

*the turtle is a reptile*

*the eagle is a bird*

*the dog is a mammal*

*...*

(all lowercase, without punctuations) mixed to other phrases, as for instance

*fish -es live in the water*

*reptile -s have cold blood*

*the turtle is slow*

*…*

The order in which the phrases are submitted is not important. The teacher could now ask the system to tell him an animal belonging to one of the categories that he used before, e.g.

*tell me a mammal*

Clearly at this point the system has no idea of the meaning of this phrase, because it started from a blank condition (*tabula rasa*). However, it can use this phrase to start an exploration phase, during which the system can retrieve phrases memorized by the association mechanism and build new phrases through partially-random action sequences. The teacher can suggest to the system a target phrase or a target word group. The exploration process is terminated when it produces the target phrase / target word group, or when the number of iterations becomes greater than a predefined limit.

For instance, if the teacher types the command:

   *.word_group mammal*

the system starts an exploration phase, which terminates when the target word group "mammal" is extracted from the working phrase buffer. Therefore, the command:

   *.phrase the dog is a mammal*

starts another exploration phase that is terminated when the working phrase, which is retrieved from the word group through the association mechanism, becomes equal to the target phrase. At this point, the teacher can type the command:

   *.word_group dog*

The system will start a new exploration phase, which terminates when the target word group "dog" is extracted from the working phrase buffer. The word group corresponds to a good output, so the teacher can reward the system using the command

   *.reward*

During the reward phase, the system retrieves the state-action sequence that led to a good output. The association between each state of the sequence and each corresponding action is rewarded by changing the connection weights of the state-action association SSM through the DHL rule.

Finally, the teacher can ask the system to say an animal belonging to a category different from the one used for training it, e.g.

   *tell me a reptile*

and start the exploitation operating mode through the command

   *.exploitation*

At the end of the exploitation phase, the system will respond with a correct output. If the question is repeated, the system will answer with another correct answer, generally different from the previous one.

This test, as well as several other tests used in the cross validation, show that the ANNABELL system is able to learn that the "is a" couple is used in phrases as "the dog is a mammal" to state that a concept belongs to a category, and that the "tell me a" group in a question can be used for asking to retrieve a concept from a category. There are two key features of the ANNABELL system that are particularly important for this type of abstraction.

- The connections that are affected by the reward (i.e. the connections of the state-action association SSM) are connected to action neurons, rather than being directly connected to output words or phrases. In this way, the system learns preferentially to build the output through sequences of elementary operations on word groups or phrases.
- The input of the state-action association SSM includes equal-words vectors. In the previous example, when the system retrieves the phrases associated to the word group "a reptile", it will recognize that the phrase "the lizard is a reptile" is more appropriate than other similar phrases (as for instance "the raven is a bird") because the last word ("reptile") is equal to the second word of the word group and to the last word of the input phrase ("tell me a reptile"), as in the association used in the training example ("a mammal" with "the dog is a mammal").

# Appendix B    Example from CHILDES database

From CHILDES database, Warren-Leubecker corpus, file david.cha.

## B.1    Comparison between ANNABELL system (left) and real child (right)

| ANNABELL system | Real child |
|---|---|
| Participants: | Participants: |
| - SYS, Ann, ANNABELL system; | - CHI, David, target child, 5 years 10 months; |
| - TEA, Teacher. | - MOT, Mother. |

| ANNABELL system | Real child |
|---|---|
| *TEA:  Ann, why do not you tell, tell Dad about you playing Pac_Man . | *MOT:  Christopher , why don't you tell , tell Daddy about you playing Pac_Man . |
| *SYS:  Michael brought me to the bumpers junction . | *MOT:  didn't Michael take you to play Pac_Man at bumper's junction ? |
| *SYS:  I played Pac_Man . | *MOT:  huh ? |
| *TEA:  did you win ? | *CHI:  yeh . |
| *SYS:  yes, I won . | *MOT:  well ? |
| *TEA:  did you play other games ? | *MOT:  did you win ? |
| *SYS:  I played Space_Invaders . | *MOT:  what did you do ? |
| *TEA:  did you win ? | *CHI:  I won ! |
| *SYS:  no, I did not win . | *MOT:  how many games did you play ? |
| *TEA:  how many games did you play ? | *CHI:  &um , I don't know . |
| *SYS:  I played : | *MOT:  you don't know ? |
| *SYS:  Space_Invaders, one ; | *MOT:  I know what's better to talk about . |
| *SYS:  Pac_Man, two ; | *MOT:  you need to tell (th)em about , &um , junk+man . |
| *SYS:  Donkey_Kong, three ; | *CHI:  xxx . |
| *SYS:  three games . | *MOT:  your letter man ! |
| *TEA:  I know what is better to talk about . | *CHI:  no ! |
| *TEA:  you need to tell them about jumble junkman . | *MOT:  yes ! |
| *SYS:  jumble junkman will be my letter man . | *MOT:  tell (th)em what I put all over your shirt . |
| *TEA:  how about Harry hair ? | *MOT:  &um &um . |
| *SYS:  Harry hair will be Bob's letter man . | *CHI:  sticker . |
| *TEA:  tell me what Mum put on your shirt . | *MOT:  sticker . |
| *SYS:  a sticker . | *CHI:  and a dress for a Barbie doll , and a wheel , and part of a necklace , and a shoestring , and a ring , or whatever . |
| *TEA:  a sticker . | *MOT:  a bracelet . |
| *SYS:  a dress_for_doll . | *CHI:  a bracelet . |
| *TEA:  ok . | *CHI:  a key , a coat+hanger . |
| *SYS:  a wheel . | *MOT:  and what's on the back ? |
| *TEA:  right . | *CHI:  &um . |
| *SYS:  a necklace . | *MOT:  a piece of paper . |
| *TEA:  a necklace . | *CHI:  piece of paper . |
| *SYS:  a shoestring . | *CHI:  and i@l , and a j@l , and a pencil , broken , and a button , and a chain , and a ring , and a crayon , and a peanut , two j@l-s +... |
| *TEA:  a shoestring . | *MOT:  how about those safety+pins ? |
| *SYS:  a ring . | *CHI:  I don't see (th)em . |
| *TEA:  a ring . | *MOT:  all those safety pins that are holding the things on your shirt . |
| *SYS:  a bracelet . | *CHI:  safety pin . |
| *TEA:  right . | *CHI:  string , string , string , string . |
| *SYS:  a j@l . | |
| *TEA:  what kind of j@l was that ? | |
| *SYS:  a capital j@l . | |

| | | | | |
|---|---|---|---|---|
| *TEA: | how about those safety_pins ? | | *MOT: | and what kind of j@l is that ? |
| *SYS: | Mum used some safety_pins to hold the things on the shirt . | | *CHI: | capital . |
| *TEA: | so, when will you wear your letter man dress ? | | *MOT: | and what kind is that ? |
| | | | *CHI: | a lower case . |
| *SYS: | May the twelfth . | | *MOT: | a lower case , that's right . |
| *TEA: | I think you wear it May the third ? | | *MOT: | so when are you gonna wear that to school ? |
| *SYS: | no May the twelfth . | | *MOT: | May the twelveth [: twelfth] , I think . |
| *TEA: | May the twelfth ? | | *CHI: | I think we wear it May the third . |
| *SYS: | yes . | | *MOT: | no , it's May the twelveth [: twelfth] . |
| *TEA: | tell me the names of the letter men . | | *MOT: | is everybody in class gonna be wearing something ? |
| *SYS: | a@l achoo . | | *CHI: | mhm . |
| *TEA: | achoo ? | | *MOT: | well what are the other lettermen , letter people that there's gonna be ? |
| *TEA: | and what, how about b@l ? | | | |
| *SYS: | b@l beautiful buttons . | | *MOT: | tell me some of the names of the letter people . |
| *TEA: | and tell me the rest of them . | | *CHI: | a@l . |
| *SYS: | c@l cotton candy . | | *MOT: | and what's a@l stand for ? |
| *TEA: | okay . | | *CHI: | achoo . |
| *SYS: | d@l delicious donuts . | | *MOT: | achoo ? |
| *TEA: | oh, I like that . | | *MOT: | and what , how (a)bout b@l ? |
| *TEA: | okay . | | *CHI: | beautiful buttons . |
| *SYS: | e@l exercise . | | *MOT: | and , tell me the rest of them . |
| *TEA: | right . | | *CHI: | c@l is cotton candy . |
| *SYS: | f@l funny feet . | | *CHI: | d@l is delicious donuts . |
| *TEA: | oh, like mine . | | *MOT: | oh I like that . |
| *SYS: | g@l gooey gum . | | *MOT: | and how about c@l ? |
| *TEA: | very good . | | *CHI: | I already told you about c@l ! |
| *SYS: | h@l Harry hair . | | *MOT: | you did ? |
| *TEA: | right . | | *CHI: | &uh huh , cotton candy . |
| *SYS: | i@l itchy itch . | | *MOT: | oh , that's right . |
| *TEA: | okay . | | *CHI: | and e@l , exercise . |
| *SYS: | j@l jumble junkman . | | *MOT: | okay . |
| | | | *CHI: | and +... |
| | | | *MOT: | f@l . |
| | | | *CHI: | f@l , funny feet . |
| | | | *MOT: | oh ! |
| | | | *MOT: | like mine . |
| | | | *CHI: | but they're big ! |
| | | | *MOT: | oh , okay . |
| | | | *CHI: | and +... |
| | | | *MOT: | g@l . |
| | | | *CHI: | gooey gum . |
| | | | *MOT: | gooey gum , alright . |
| | | | *CHI: | and , h@l , Harry hair . |
| | | | *MOT: | I . |
| | | | *CHI: | scratchy scratch . |
| | | | *MOT: | no , that's not for i@l . |
| | | | *MOT: | is that itchy something ? |
| | | | *CHI: | itchy itch . |
| | | | *CHI: | and +... |
| | | | *MOT: | and you got j@l , and what'd you say that was ? |
| | | | *CHI: | jumble junk+man . |
| | | | *MOT: | jumble junk+man . |

## B.2 Informative sentences used to build the experience of the system in a text-based virtual environment

########################################################
you are at the bumpers junction
Michael brought you here
you play Space_Invaders
you do not win
you play Pac_Man
you win
you play Donkey_Kong
you do not win
you leave

########################################################
you are at school
you can see the teacher your classmate -s
# the teacher says
next week we will put on a show about the letter -s of the alphabet
each of you will play a letter man
your mum will help you to prepare the dress for your letter man
you will wear your letter man dress May the twelfth
the letter men are
a@l achoo
b@l beautiful buttons
c@l cotton candy
d@l delicious donuts
e@l exercise
f@l funny feet
g@l gooey gum
h@l Harry hair
i@l itchy itch
j@l jumble junkman
Susan your letter man will be the achoo
Antony your letter man will be beautiful buttons
Andy your letter man will be cotton candy
Nicole your letter man will be delicious donuts
Carol your letter man will be exercise
Helen your letter man will be funny feet
Matt your letter man will be gooey gum
Bob your letter man will be Harry hair
Chris your letter man will be itchy itch
Ann your letter man will be jumble junkman

####################################################
you are at home
you can see Mum
#Mum says
we have to prepare the dress for your letter man jumble junkman
we can put many thing -s all over your shirt
#
Mum put -s a sticker on your shirt
Mum put -s a dress_for_doll on your shirt
Mum put -s a wheel on your shirt
Mum put -s a necklace on your shirt
Mum put -s a shoestring on your shirt
Mum put -s a ring on your shirt
Mum put -s a bracelet on your shirt
Mum put -s a j@l on your shirt
Mum has done put -ing things
the j@l is a capital j@l
Mum use -s some safety_pin -s to hold the thing -s on the shirt

####################################################
you are at the central park
Dad brought you here
you play hide_and_seek
you do not win
Susan win -s
you play bowl -s
you win
you play frisbee
you do not win
Oliver win -s
you play tag
you win
you leave

####################################################
you are at school
you can see the teacher your classmate -s
# the teacher says
Ann your desk mate will be Susan Dale
Matt your desk mate will be Bob Fingers
Helen your desk mate will be Carol Davis

####################################################
you are in Susan 's room
you can see Susan aunt Carol

\# aunt carol says  
Susan you must tidy up your room  
\#  
Susan put -s a doll in her toy_box  
Susan put -s a doll_house in her toy_box  
Susan put -s a doll_dress in her toy_box  
Susan put -s a toy_horse in her toy_box  
Susan put -s a toy_dog in her toy_box  
Susan put -s a puppet in her toy_box  
Susan has done put -ing things  
the puppet is a white cat  
Susan place -s some book -s in the bookshelf  

######################################################  
Ann you will celebrate your fifth birthday February the seventh  

######################################################  
the fantastic four are  
Mister Fantastic  
\#Reed Richards  
The Invisible Woman  
\#Susan "Sue" Storm  
The Thing  
\#Ben Grimm  
The Human Torch  
\#Johnny Storm  

######################################################  
the seven kings are  
first Romulus  
second Numa Pompilius  
third Tullus Hostilius  
fourth Ancus Marcius  
fifth Tarquinius Priscus  
sixth Tarquinius Superbus  
seventh Servius Tullius  

## B.3   Training sentences

*TEA: tell me about you play -ing bowl -s  
*SYS: Dad brought me to the central park  
*SYS: I play -ed bowl -s  
*TEA: ? did you win  
*SYS: yes I won  

*TEA: ? did you play other game -s
*SYS: I play -ed hide_and_seek
*TEA: ? did you win
*SYS: no I did not win
*TEA: ? how many game -s did you play
*SYS: I play -ed
*SYS: hide_and_seek one
*SYS: bowl -s two
*SYS: frisbee three
*SYS: tag four
*SYS: four game -s

*TEA: tell me about Susan Dale
*SYS: Susan Dale will be my desk mate

*TEA: tell me about Bob Fingers
*SYS: Bob Fingers will be Matt's desk mate

*TEA: tell me what Susan put in her toy_box
*SYS: a doll
*SYS: a doll_house
*SYS: a doll_dress
*SYS: a toy_horse
*SYS: a toy_dog
*SYS: a puppet
*TEA: ? what kind of puppet was that
*SYS: a white cat
*TEA: ? how about those book -s
*SYS: Susan place -ed some book -s in the bookshelf

*TEA: Ann
*TEA: ? when will you celebrate your fifth birthday
*SYS: February the seventh
*TEA: ? February the sixth
*SYS: no February the seventh
*TEA: ? February the seventh
*SYS: yes

*TEA: tell me the name -s of the fantastic four
*SYS: Mister Fantastic
*TEA: Mister Fantastic
*SYS: The Invisible Woman
*TEA: right
*SYS: The Thing
*TEA: that is right

\*SYS: The Human Torch

\*TEA: tell me the name -s of the seven kings
\*SYS: first Romulus
\*TEA: first Romulus
\*SYS: second Numa Pompilius
\*TEA: second Numa Pompilius
\*SYS: third Tullus Hostilius
\*TEA: that is good
\*SYS: fourth Ancus Marcius
\*TEA: right
\*SYS: fifth Tarquinius Priscus
\*TEA: very good
\*SYS: sixth Tarquinius Superbus
\*TEA: okay
\*SYS: seventh Servius Tullius

## Appendix C    Neural activation patterns in a concrete example

This section describes through a concrete example how the neural activation patterns of the system evolve, how the connection weights are modified during the training stage and how these weight changes make the system able to generalize the response to new sentences. A detailed description of the system architecture is provided in the Supporting information document *The ANNABELL system architecture*.

### C.1    Input phrase acquisition

Fig 13 illustrates how an input sentence ("the turtle is a reptile" in this example) is acquired by the system and stored in the input phrase buffer. When this sentence is written in the terminal or read from a file, the interface submits its words one by one to the system, using the ascii representation, starting from the word "the". This word is mapped to the *input-word buffer (*IW) through the mechanism described in *The ANNABELL system architecture*, Sect. 2. The input nodes are fully connected to IW, and the connection weights are initialized randomly. IW is updated using the *winner-take-all* (WTA) rule: the neuron with the highest activation state (winner neuron) is switched to the level one, while all other neurons of IW are switched to zero. The connections from the input nodes to the winner neuron of IW are updated through the *discrete-hebbian-learning* (DHL) rule: if the input node signal is one, the connection weight is saturated to its maximum value (+1), otherwise it is saturated to its minimum value (-1). This ensures that if this word is submitted again to the system, the winner neuron will be the same.

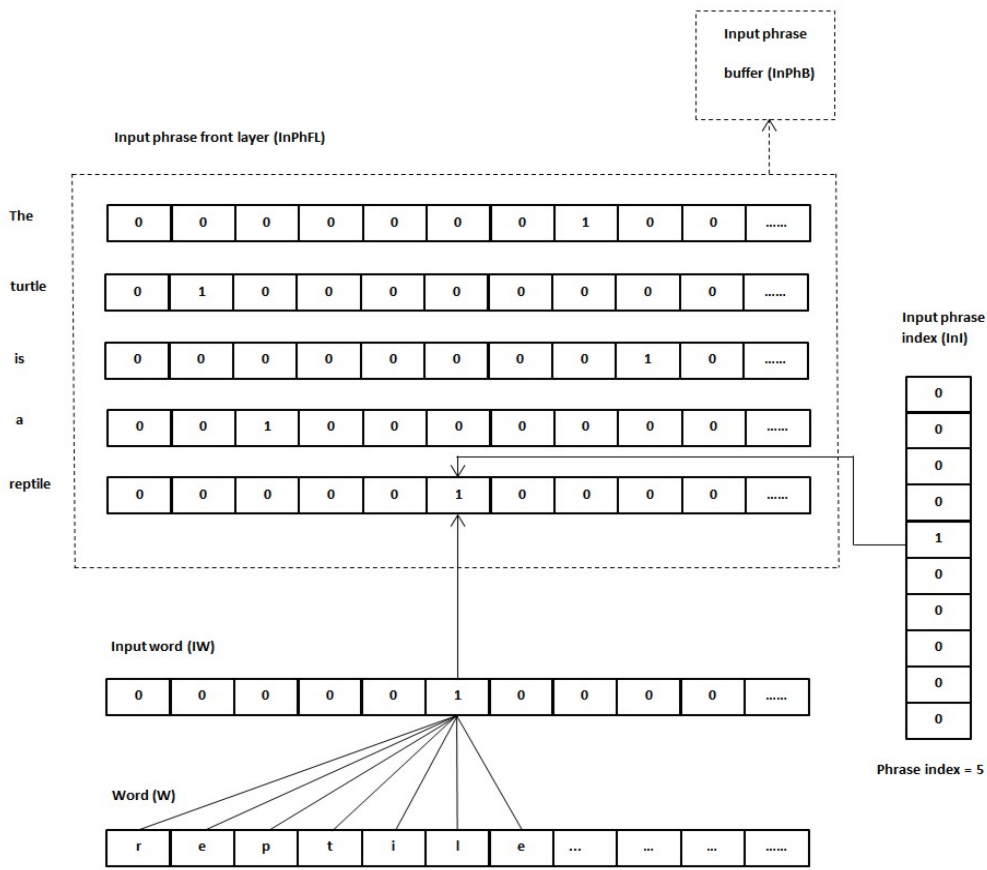

**Fig 13.** Input phrase acquisition.

PhI (*phrase index*) is a subnetwork that represents the position of the current word in the phrase: the neuron of PhI corresponding to the position of the word in the phrase is in a high-level state, while all the others are in a low-level state. The words of the input phrase are submitted to the system by loading them, one by one, in the word buffer, and increasing the phrase index from 1 to the number of words in the phrase. The system itself initializes the phrase index at the beginning of a phrase acquisition, and increases it after the acquisition of each word, as discussed in *The ANNABELL system architecture*, Sect. 9. In this way a couple (word, phrase-index) is mapped to the neuron of InPhFL located in the row i corresponding to the phrase index and in the column j corresponding to the word-mapping neuron index.

*This structure is suitable for a broad range of problems in adaptive behavior, not only language understanding. In general, a "word" can be defined as a specific input pattern. The system can associate a key to each word received as input and generate a unique pattern corresponding to the*

*couple (key, word). A "phrase" is set of couples (key, word), temporarily stored in the system. The key can be any pattern, not necessarily representing an integer number, however in the SSM approach a single neuron or a small number of neurons should be active for any key pattern. In the case of natural language, the "phrase index" is a key that represents the position of each word in a phrase.*

The input-phrase front layer is single-connected to the input-phrase buffer (InPhB). The input-phrase buffer is also single-connected to itself (self connection). In this way, it can store all words of a phrase and keep them stored until it is cleared by a flush signal.

## C.2    Copy of the input phrase to the working-phrase buffer

After the whole input sentence is acquired, the system executes the action PH_FROM_INPUT, which copies the sentence from the input phrase buffer to the working phrase buffer, as illustrated in Fig 14. The input-phrase-buffer neurons are connected one by one to the intermediate subnetwork WkPhFL. The action neuron PH_FROM_INPUT activates the gatekeeper neuron WkFlag, which is fully connected to WkPhFL. When the gatekeeper neuron is ON, WkPhFL allows the signal to flow from the input phrase buffer to the working phrase buffer, through the mechanism described in Sect. 2.2.

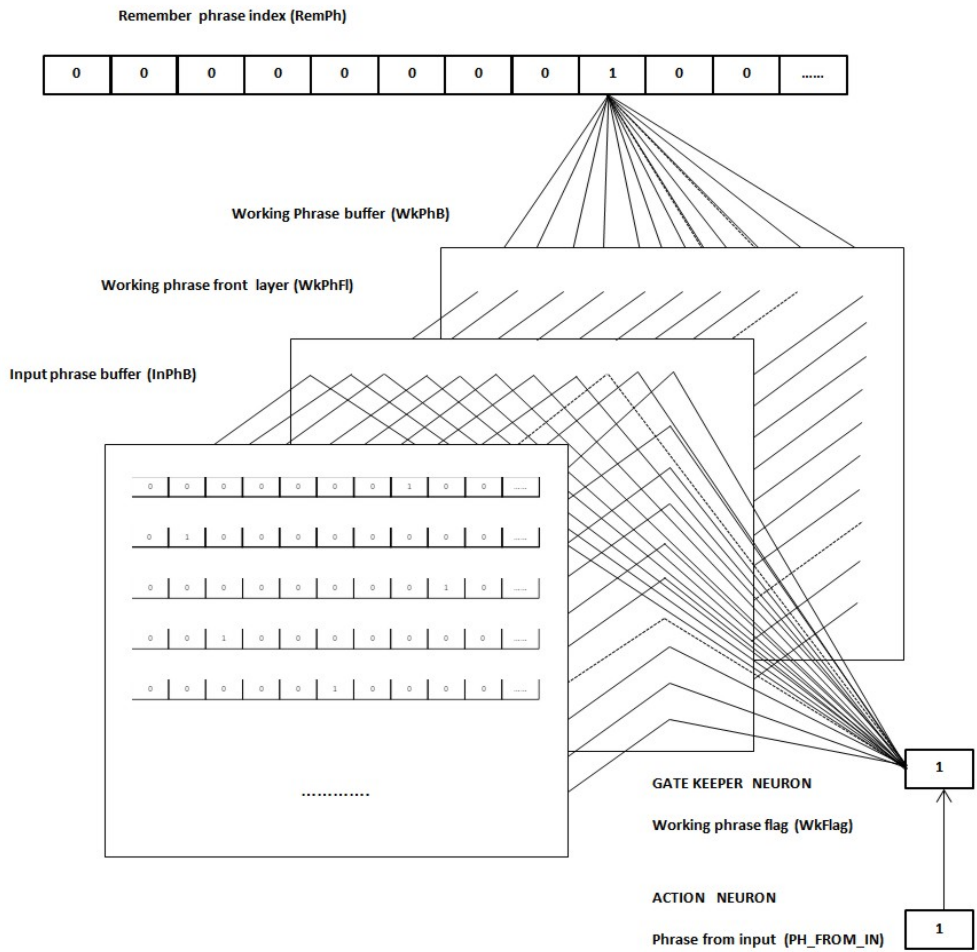

**Fig 14.** Copy of the input phrase to the working phrase buffer and memorization of the phrase in the long-term memory.

## C.3 Memorization of the input phrase in the long-term memory

The subnetwork RemPh (remembered phrase) is used as an index for storing and retrieving phrases from the long-term memory. RemPh is fully connected to WkPhB by forcing connections. The active neuron of RemPh represents the current phrase index in the long-term memory. After the input phrase is copied to WkPhB, the connections from this neuron to WkPhB are updated through the DHL rule. In this way, if this neuron is switched ON again, it will retrieve the memorized phrase by forcing the activation states of WkPhB.

## C.4 Extraction of a word-group from the working-phrase buffer

Fig 15 shows how a word is extracted from the working phrase buffer. PhI (phrase index) represents the index of the word in the phrase. Each neuron of the intermediate subnetwork WkWfI performs a logical AND between the corresponding neuron of WkPhB and that of PhI. In this way, the row of WkPhB corresponding to the phrase index is copied to the subnetwork that represents the current word, CW.

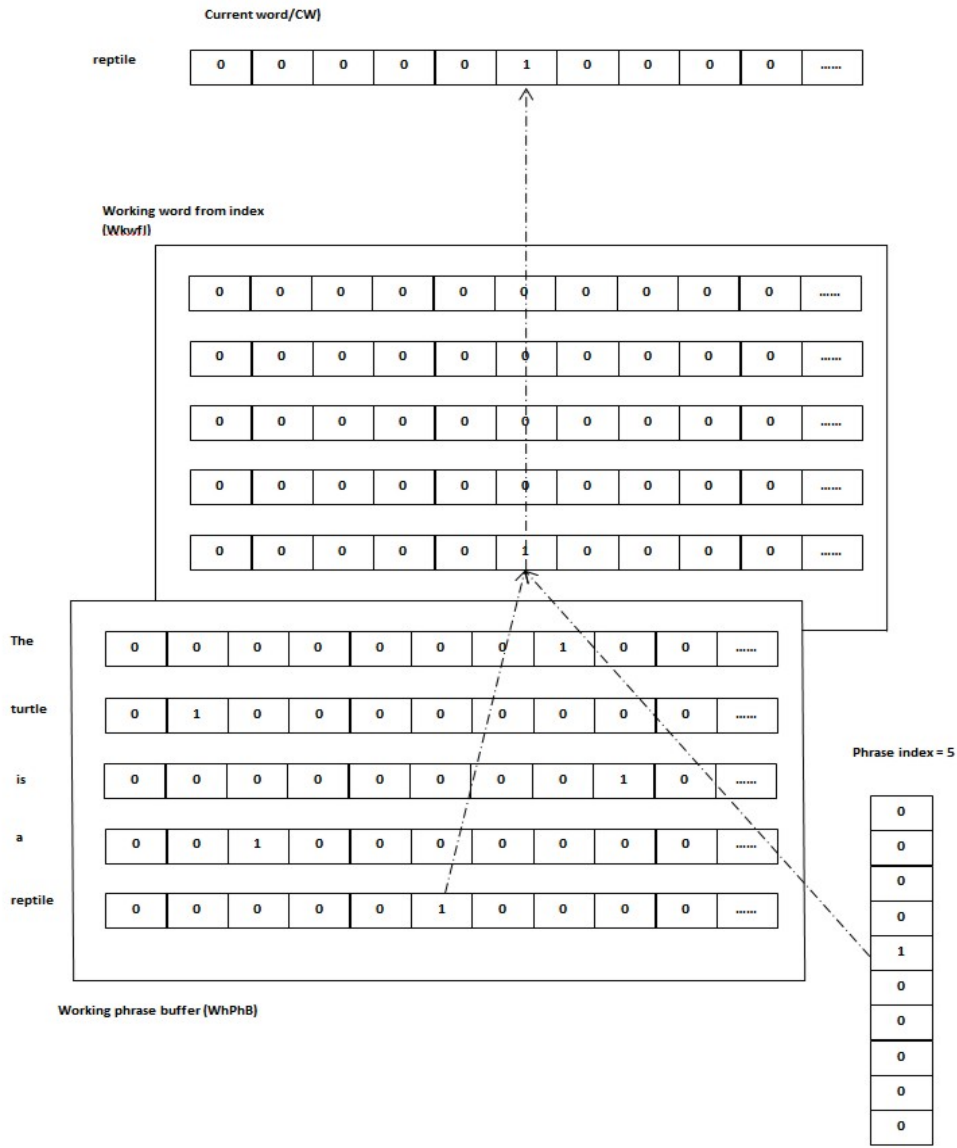

**Fig 15.** Extraction of the current word from the working phrase buffer.

The current word can be extracted from CW and copied to the word-group buffer through the procedure illustrated in Fig 16, which is controlled by the gatekeeper neuron GetFlag. When GetFlag is ON, the intermediate subnetwork WGCW (word-group corrent word) allows the flow of signal from CW to WGFL (word-group front layer), which operates a logical AND between this signal and the word-group index WGI. In this way, the current word is copied to the row of WGFL that corresponds to the word group index.

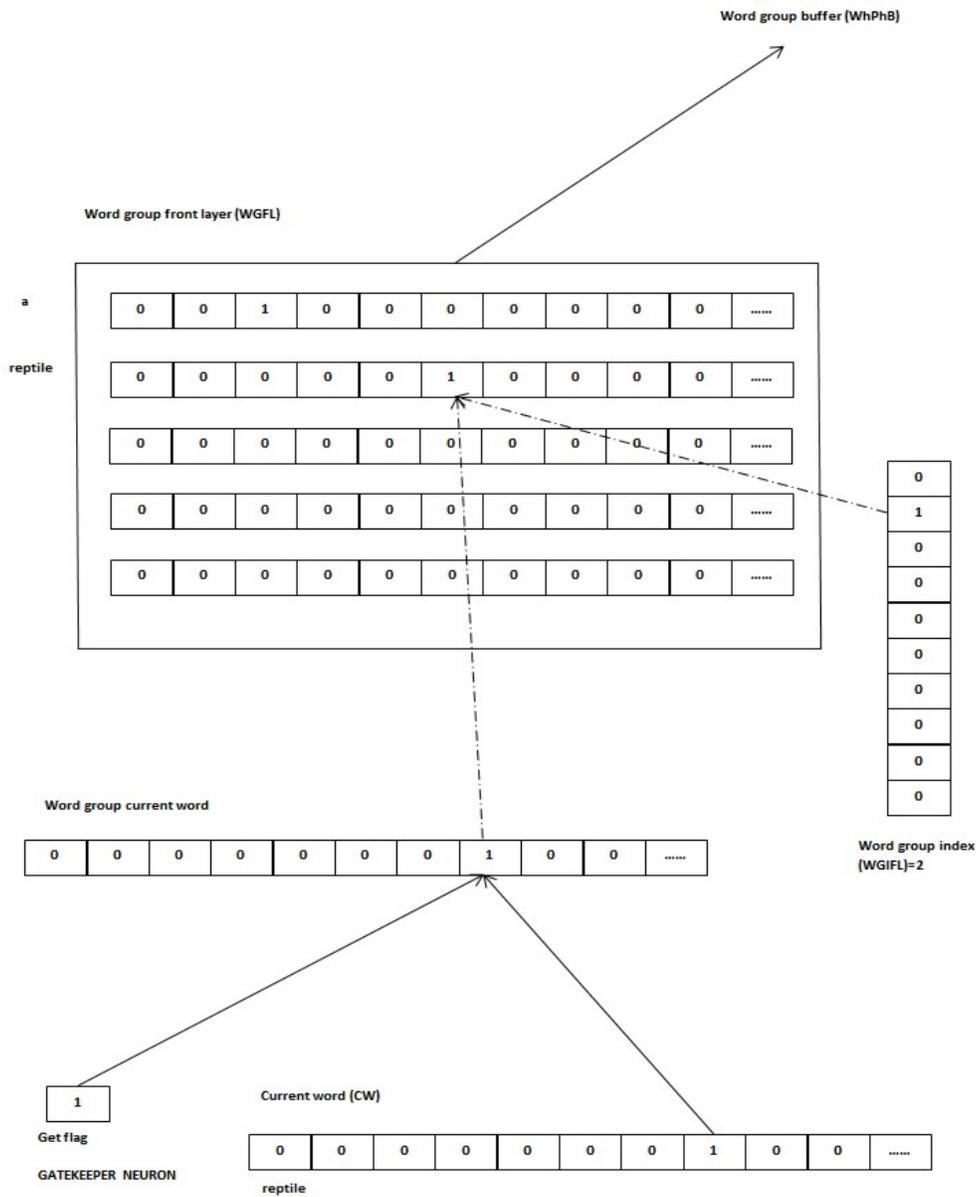

**Fig 16.** Copy of the current word to the word group buffer.

## C.5 Memorization and retrieval of the association between a word group and a phrase

The group of words in WGB can be used as a cue to retrieve a phrase from the long term memory. Fig 17 shows how the association between a group of words and the whole phrase is stored in the long-term memory. The word group buffer is fully connected to the subnetwork RemPhfWG (remembered-phrase from word group), which is fully connected to RemPh by forcing connections. RemPhfWG is updated through the WTA rule. The connections from WGB to the winner neuron and the connections from this neuron to RemPh are updated through the DHL rule.

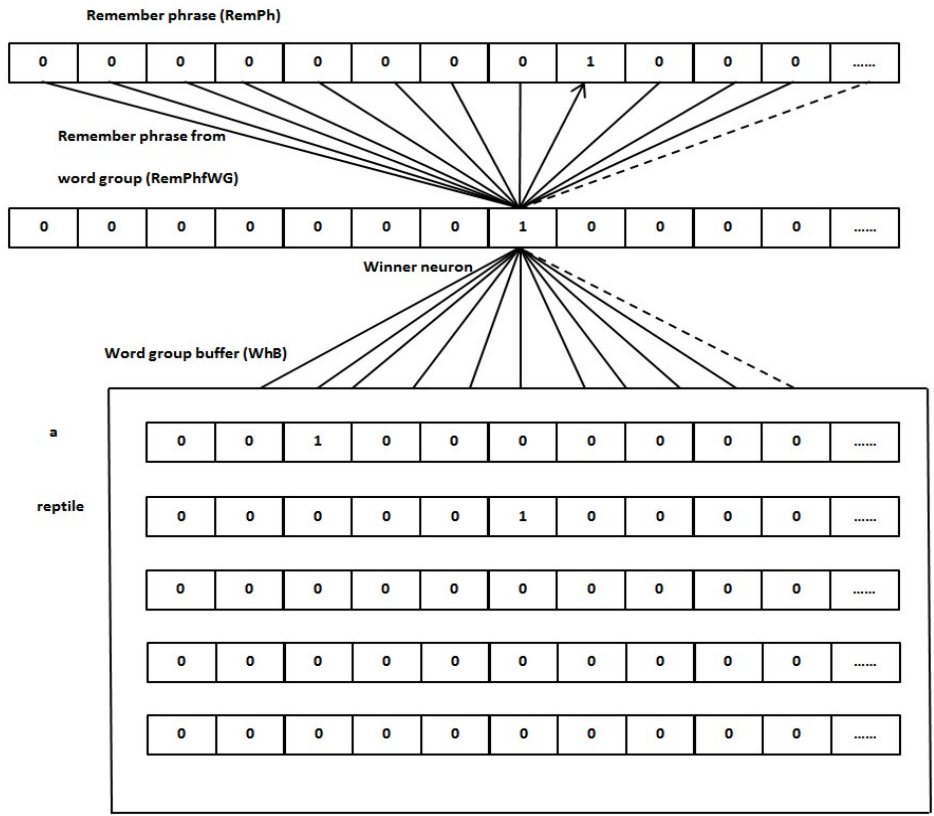

**Fig 17.** Memorization and retrieval of the association between a word group and a phrase.

The association between the word group in WGB and the phrase in WkPhB is memorized in the long-term memory using the architecture shown in Fig 17. The word group buffer is fully connected to RemPhfWG, which is fully connected to RemPh by forcing connections. RemPhfWG is updated through the *winner-take-all* (WTA) rule. The weights of the connections from WGB to the winner neuron and the weights of the connections form this neuron to RemPh are updated through the DHL rule. In this way, the association between the current content of WGB and the current content of RemPh is permanently memorized by the system. During the retrieval process, the word group in WGB is sent as input to RemPhfWG. The neurons having connection weights matching the word group will have the highest activation state, and a single winner is selected among them through the WTA rule. The winner neuron will retrieve the phrase associated to the input word group by using its forcing output connections to set the activation state of WkPhB.

## C.6 Exploration

The system is trained to respond to the input sentences through an exploration/reward procedure. Following our example, suppose that the human interlocutor submits the question-like imperative sentence

*tell me a reptile*

During the exploration phase, the system performs partially random action sequences.

The basic action sequence is that described in Sect. 2.3:

- *W_FROM_WK*
- *NEXT_W* ($N_1$ times)
- *FLUSH_WG*
- *GET_W, NEXT_W* ($N_2$ times)
- *RETR_AS*

with $N_1$, $N_2$ random integer numbers. The action neuron NEXT_W activates the gatekeeper neuron NextPhIFlag, which triggers an increase of the phrase index PhI, as described in *The ANNABELL system architecture*, Sect. 9. The action neuron GET_W activates the gatekeeper neuron GetFlag, which controls the copy of the current word from the working phrase to the word group buffer, as described previously. The action neuron RETR_AS activates the gatekeeper neuron RetrAs, which controls the retrieval of a phrase from the long-term memory using the word group as a cue as discussed in the above paragraphs.

The whole sequence is repeated, using different random integer values for $N_1$, $N_2$, until it produces the target output. In our example this can occur, for instance, with $N_1=2$ and $N_2=2$. In fact, in this case the system extracts the word group "a reptile" from the input phrase "tell me a reptile". The RETR_AS action uses this word group as a cue, and can eventually retrieve the phrase "the turtle is a reptile" from the long-term memory. The basic action sequence is repeated on the new working phrase, and the system produces the target output "turtle" if $N_1=1$ and $N_2=1$.

The state-action sequences are memorized through the mechanism represented in Fig 18. The state-action index StActI is initialized to one at the beginning of each sequence, and it is increased every time the system produces a new action. The neurons of StActI are connected one by one to the corresponding neurons of StActMem (state-action memory), which is fully connected to all subnetworks that represent the internal state of the system state (as defined in Sect. 2.3) and to the action neurons. The output connections of StActMem are updated through the DHL rule. In this way, StActMem can retrieve the state-action sequence by forcing the activation state of the neurons connected to its output connections.

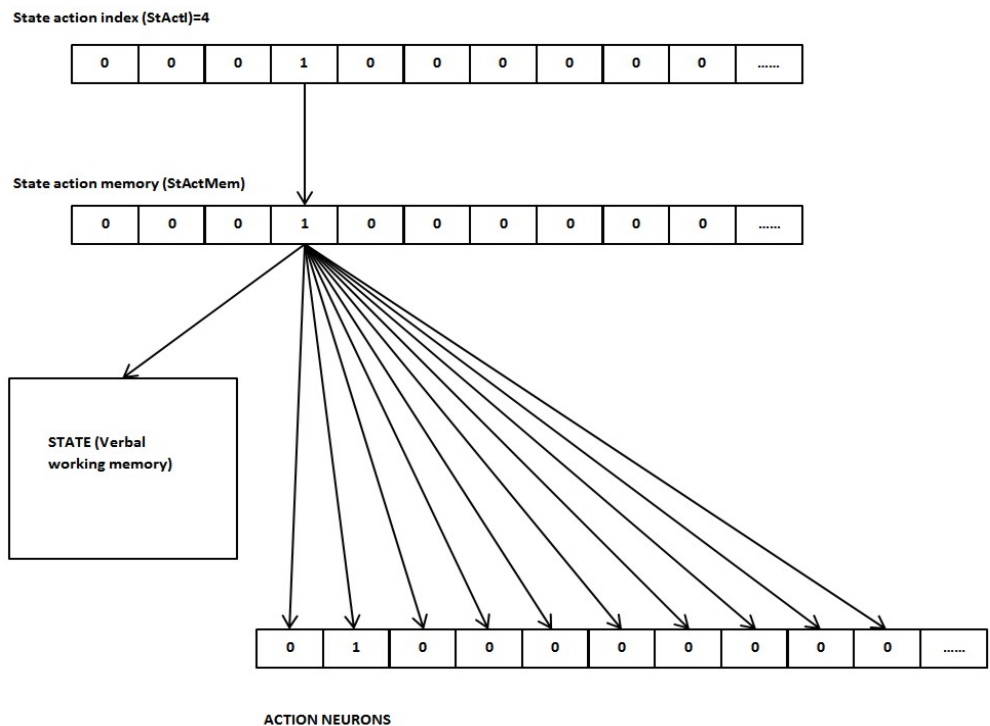

**Fig 18.** Memorization and retrieval of a state-action sequence.

## C.7 Reward

When the exploration phase leads to the target output, the system is set to the reward operating mode. The memorized state-action sequence is retrieved, as described in the previous paragraphs.

The association between each state of the sequence and the corresponding action is memorized through the state-action association subnetwork ElActfSt, which has input connections fully connected to the system state and output connections fully connected to the action neurons, as illustrated in Fig 19. In the reward operating mode, ElActfSt is updated through the WTA rule, and both the connections from the verbal working memory to the winner neuron of ElActfSt and the connections from this neuron to the action neurons are updated through the DHL rule.

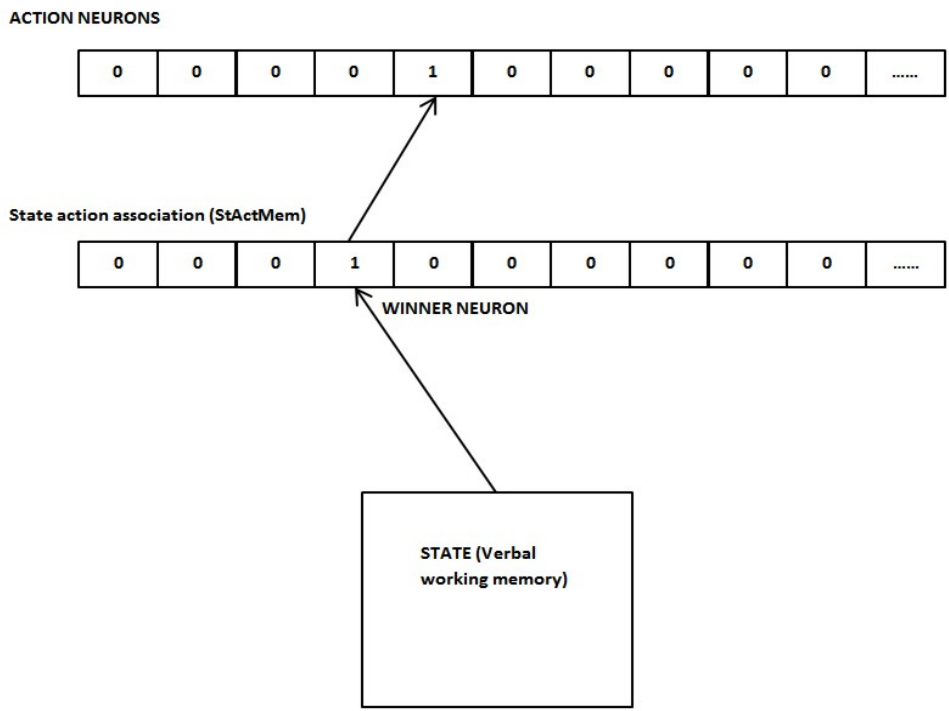

**Fig 19.** State-action association.

## C.8 Exploitation

Following our example, suppose that the human interlocutor types the sentence

*tell me a mammal*

and that after the acquisition the system is set to the exploitation operating mode.

The state-action association subnetwork ElActfSt receives its input from the system state, and it is updated through the k-WTA rule: the k neurons that have the highest activation state are set to one, while all other neurons are set to zero. Those neurons send their output to the action neurons, which are updated through the WTA rule: the action neuron with the highest activation state (which represents the action with the highest score) is set to one, while all other action neurons are set to zero.

Although the input phrase is new for the system, it is similar to the one used for training ("tell me a reptile"). Therefore, at each step of the exploitation phase, the neurons of the central executive with the highest activation state will be those that have been rewarded in the training example, and consequently the action sequence will be the same, i.e.

- *W_FROM_WK*
- *NEXT_W* (2 times)
- *FLUSH_WG*
- *GET_W, NEXT_W* (2 times)
- *RETR_AS*
- *W_FROM_WK*
- *NEXT_W* (1 time)
- *FLUSH_WG*
- *GET_W, NEXT_W* (1 time)

Through such sequence, the system will extract the word group "a mammal" from the working phrase, retrieve a phrase as "the dog is a mammal" from the long term memory, extract the word group "dog" and send it to the output.

## Appendix D   Mathematical properties of the state-action association system

The heart of the ANNABELL model is the state-action association system, which is responsible for all decision processes, as described in Sect. 2.3. This system is implemented as a neural network (state-action association neural network, abbreviated as SAANN) with input connections fully connected to all subnetworks of the short-term memory (STM), which represents the internal state of the system, and output connections fully connected to the set of mental action neurons. Therefore, the SAANN receives as input the internal state and yields as output a mental action. The input and output connections of this system have learnable weights, which are updated through a discrete version of the Hebbian learning rule (DHL rule). Furthermore, the activation states of the SAANN are updated through a variant of the k-winner-take-all rule, while those of the action neurons are updated through the (one-) winner-take-all rule.

In this section, we describe the update rules in more details and we prove that our model of the state-action association system is equivalent to a k-nearest-neighbor (k-NN) classifier with a proper definition of the distance in the input space. For large enough training sets, the k-NN algorithm is guaranteed to yield an error rate no worse than twice the Bayes error rate, which is the minimum achievable error rate given the distribution of the data [51].

The discrete-Hebbian-learning (DHL) rule used in our model is an extreme simplification compared with other models more focused on biological realism. The same type of simplification is often used in neural models of memory based on the Hopfield recurrent neural networks [37].

Nessler et al. [52] have proven that a more realistic model of Hebbian learning, in combination with a sparse neural code, can learn to infer optimal Bayesian decisions for arbitrarily complex probability distributions. However, a more realistic implementation of the Hebbian learning rule, with small updates of the connection weights, would require very large computational resources for training and evaluating our model on large datasets, and real time interaction with the system would not be possible. O'Reilly [38] have shown that the k-winner-take-all rule is biologically justified.

Other simplifications are used in our model:
- *stability condition*: the proof that the state-action association system is equivalent to a k-NN classifier assumes that the STM can be partitioned into $M$ subnetworks each having a fixed number of neurons active at a time. The weight saturation value $W_{max}$, used by the DHL rule, is assumed to be the same for all connections of the same subnetwork. A particular case is when the whole STM has a fixed number of neurons active at a time, and $W_{max}$ has the same value for all connections from the STM to the SAANN. In the subnetworks that represent words or phrases, the stability condition is ensured by using default neurons, which represent the null word. In the subnetworks used for word comparison, such property is fulfilled by representing the two conditions, equal/not-equal word, using two complementary neurons instead of one.

- During the training stage, the SAANN is updated through the new-winner-take-all rule: a previously unused neuron is set to the high-level activation state ("on"), while all other neurons are set to the low-level activation state ("off").

Those two simplifications are used only for ensuring validity of the k-NN equivalence theorem, so that the statistical properties of the model are contextualized in a well-known theoretical framework, and good convergence properties of the error rate are guaranteed.

It is worth to mention that many biologically inspired neural models of language use the standard backpropagation learning algorithm, even though it does not have a biological justification, because it ensures error minimization. In contrast, our model is based on the same learning principle that is responsible for synaptic plasticity in biological neural networks.

Let $A_m$ and $W_{\max,m}$ be the number of active neurons and the weight saturation value of the $m^{th}$ subnetwork, respectively. The stability condition ensures that $A_m$ is constant. Let $s_{mj}$ be the activation state (0 or 1) of the $j^{th}$ neuron of the $m^{th}$ subnetwork. The sum and the square sum of $s_{mj}$ weighted with $W_{\max,m}$ are the following:

$$\sum_{m=1}^{M} W_{\max,m} \sum_{j=1}^{N_m} s_{mj} \quad \text{and} \quad \sum_{m=1}^{M} W_{\max,m} \sum_{j=1}^{N_m} s_{mj}^2 \qquad (1)$$

where $N_m$ is the number of neurons of the $m^{th}$ subnetwork. The stability condition implies that, for all values of $m$,

$$\sum_{j=1}^{N_m} s_{mj} = \sum_{j=1}^{N_m} s_{mj}^2 = A_m \qquad (2)$$

therefore the following *normalization conditions* can be derived for the weighted sum and for the weighted square sum of the signal:

$$\sum_{m=1}^{M} W_{\max,m} \sum_{j=1}^{N_m} s_{mj} = U_1 \quad \text{and} \quad \sum_{m=1}^{M} W_{\max,m} \sum_{j=1}^{N_m} s_{mj}^2 = U_2 \qquad (3)$$

where

$$U_1 = U_2 = \sum_{m=1}^{M} W_{\max,m} A_m \qquad (4)$$

are constants. It is worth to point out that these two normalization conditions are sufficient for the validity of the k-NN equivalence theorem, which we will prove below, even if the stability condition is not satisfied.

The weighted distance between two states $S_1$ and $S_2$ of the STM can be defined as:

$$d(S_1, S_2) = \sum_{m=1}^{M} W_{max,m} \sum_{j=1}^{N_m} (s_{1mj} - s_{2mj})^2 = 2U_2 - 2\sum_{m=1}^{M} W_{max,m} \sum_{j=1}^{N_m} s_{1mj} s_{2mj} \qquad (5)$$

where we used the second normalization condition of Eq. 3.

Let $N_A$ be the number of mental action neurons, i.e. the number of possible actions that can be triggered by the state-action association system. A mental action can be represented by an integer value:

$$a = 1, \ldots, N_A \qquad (6)$$

A state-action sequence, starting with a state $S^\alpha_1$ and ending in a state $S^\alpha_{T\alpha}$ will be called an epoch:

$$(S^\alpha_1, a^\alpha_1), \ldots, (S^\alpha_t, a^\alpha_t), \ldots, (S^\alpha_{T\alpha}, a^\alpha_{T\alpha}) \quad . \qquad (7)$$

The index $\alpha$ represents the epoch, while the index $t$ represents the time step in the epoch:

$$t = 1, \ldots, T_\alpha \quad . \qquad (8)$$

The number of time steps in all epochs is limited: $T_\alpha \leq T_{max}$. An epoch can receive a reward depending only on its final state. In the reward phase, the state-action memory retrieves the whole state-action sequence, and the SAANN is trained using the state $S^\alpha_t$ as input and the corresponding action $a^\alpha_t$ as target output. At each time step $t$ of the sequence, the SAANN is updated using the new-winner-take-all rule: a previously unused neuron $i$ is set to the "on" state, while all other neurons are set to the "off" state. The connections from the STM to the winner neuron are updated through the DHL rule:

$$w_{imj} = \begin{cases} +W_{max,m} & \text{for } s^\alpha_{tmj} = 1 \\ -W_{max,m} & \text{for } s^\alpha_{tmj} = 0 \end{cases} \qquad (9)$$

where the two indexes $m$ and $j$ refer to the $j^{th}$ neuron of the $m^{th}$ subnetwork of the STM, $s^\alpha_{tmj}$ is the activation state of this neuron (0 or 1) at the epoch $\alpha$ and time step $t$, $w_{imj}$ is the weight of the connection to the $i^{th}$ neuron of the SAANN (i.e. the winner neuron) and $W_{max,m}$ is the weight-saturation absolute value for the subnetwork $m$. These two equations can also be written as:

$$w_{imj} = W_{max,m}(2 s^\alpha_{tmj} - 1) \qquad (10)$$

The connections from the winner neuron of the SAANN to the action neurons are also updated through the DHL rule:

$$w_{li} = \begin{cases} +1 & \text{for } l = a_t^\alpha \\ -1 & \text{for } l \neq a_t^\alpha \end{cases} \quad (11)$$

where the index $l$ refers to an action neuron, $w_{li}$ is the weight of the connection from the winner neuron of the SAANN to this action neuron and $a^\alpha_t$ is the target action.

During the exploitation phase, in general the internal states will be different from those used in the training phase. Let $S^{test}$ be a generic internal state of the system in the exploitation phase. The total input signal to each neuron of the SAANN is:

$$y_i = \sum_{m=1}^{M} \sum_{j=1}^{N_m} w_{imj} s_{mj}^{test} \quad (12)$$

where the bias signal is assumed to be null. From Eq. 10 for $w_{imj}$ it follows that

$$y_i = \sum_{m=1}^{M} [W_{max,m} \sum_{j=1}^{N_m} (2 s_{tmj}^\alpha - 1) s_{mj}^{test}] = 2 \sum_{m=1}^{M} W_{max,m} \sum_{j=1}^{N_m} s_{tmj}^\alpha s_{mj}^{test} - U_1 \quad (13).$$

where we used the first normalization condition of Eq. 3, and using Eq. 5 for the weighted distance:

$$y_i = 2 U_2 - U_1 - d(S_t^\alpha, S^{test}) \quad (14)$$

In the exploitation phase, the SAANN is updated through the k-winner-take-all rule: the k neurons with the highest activation state are set "on", while all the others are set "off".

Since the activation function $f(y_i)$ is an increasing function of the input signal $y_i$, from Eq. 14 it follows that the neurons with the highest activation state $y_i$ are those with the smallest value of $d(S_t^\alpha, S^{test})$. Therefore, the $k$ neurons with the highest activations are those that correspond to the training internal states that have the smallest weighted distance from the current (test) internal state, i.e. to the $k$ nearest neighbors with such metric.

Each "used" neurons of the SAANN, i.e. each neuron that was classified as a winner neuron during a reward phase, is connected with a positive-weight connection ( $w_{li} = +1$ ) to one and only one action neuron, while it is connected to all other action neurons by negative-weight connections ( $w_{li} = -1$ ). We can therefore partition the used neurons of the SAANN in classes, based on the action that they "suggest".

The input signal to each action neuron is equal to the weighted sum of the input from the $k$ winner neurons. Since the output of the winner neurons is 1, and the weights are $w_{li} = \pm 1$, the input signal is equal to the number of winner neurons that "suggest" that action as the best action, minus the number of those that do not. The action neuron with the highest input signal is the one that is "suggested" as the best action by the greatest number of winner neurons. The actions neurons are updated by the (one) winner-take-all rule, therefore this neuron will be set "on", while all the other action neurons will be set "off". This is equivalent to a k-NN classification. In fact, in k-NN classification an entry is assigned to the class most common among its *k* nearest neighbors.